\documentclass[journal,comsoc]{IEEEtran}
\usepackage[T1]{fontenc}

\ifCLASSINFOpdf
\else
\fi
\usepackage{cite}
\usepackage{amsmath,amssymb,amsfonts}
\usepackage{algorithm,algorithmic}
\usepackage{graphicx}
\usepackage{xcolor}
\usepackage{algorithmic}
\usepackage{graphicx}
\usepackage{caption}
\usepackage{amsmath}

\usepackage{tabu}
\usepackage{lineno}
\usepackage{amsthm}
\usepackage{mathtools}
\usepackage{footnote}

\usepackage{subfigure}

\DeclareMathAlphabet{\mathcal}{OMS}{cmsy}{m}{n}

\DeclareMathOperator*{\argmax}{argmax}

\usepackage[cmintegrals]{newtxmath}
\begin{document}
	\title{
	Neuro-symbolic Explainable Artificial Intelligence Twin for Zero-touch IoE in Wireless Network
		}

	\author{Md.~Shirajum~Munir,~\IEEEmembership{Member,~IEEE,}
		Ki~Tae~Kim,~\IEEEmembership{}
		Apurba~Adhikary,~\IEEEmembership{}
		Walid~Saad,~\IEEEmembership{Fellow,~IEEE,}
		Sachin~Shetty, \IEEEmembership{Senior~Member,~IEEE,}
		Seong-Bae~Park,~\IEEEmembership{}
		and~Choong~Seon~Hong,~\IEEEmembership{Senior~Member,~IEEE}
		\thanks{Md. Shirajum Munir is with the Virginia Modeling, Analysis, and Simulation Center, Department of Computational Modeling and Simulation Engineering, Old Dominion University, Suffolk, VA 23435, USA, and also with the Department of Computer Science and Engineering, Kyung Hee University, Yongin-si 17104, Republic of Korea (e-mail: munir@khu.ac.kr).}
		\thanks{Ki Tae Kim, Apurba Adhikary, Seong-Bae Park, and Choong Seon Hong are with the Department of Computer Science and Engineering, School of Computing, Kyung Hee University, Kyung Hee University, Yongin-si 17104, Republic of Korea (e-mail: glideslope@khu.ac.kr; apurba@khu.ac.kr; sbpark71@khu.ac.kr; cshong@khu.ac.kr).}
		\thanks{Walid Saad is with the Wireless@VT Group, Bradley Department of Electrical and Computer Engineering, Virginia Tech, Arlington, VA 22203 USA (e-mail: walids@vt.edu).}
		\thanks{Sachin~Shetty is with the Virginia Modeling, Analysis, and Simulation Center, Department of Computational Modeling and Simulation Engineering, Old Dominion University, Suffolk, VA 23435, USA (e-mail: sshetty@odu.edu)}
		\thanks{Corresponding author: Choong Seon Hong (e-mail: cshong@khu.ac.kr)}} 

	\markboth{XXX,~Vol.~xx, No.~xx, October~2022}%
	{Shell \MakeLowercase{\textit{et al.}}: Bare Demo of IEEEtran.cls for IEEE Communications Society Journals}
	%
	
	
	
	\maketitle
	
\begin{abstract}
Explainable artificial intelligence (XAI) twin systems will be a fundamental enabler of zero-touch network and service management (ZSM) for sixth-generation (6G) wireless networks. A reliable XAI twin system for ZSM requires two composites: an extreme analytical ability for discretizing the physical behavior of the Internet of Everything (IoE) and rigorous methods for characterizing the reasoning of such behavior. 
In this paper, a novel neuro-symbolic explainable artificial intelligence twin framework is proposed to enable trustworthy ZSM for a wireless IoE. The physical space of the XAI twin executes a neural-network driven multivariate regression to capture the time-dependent wireless IoE environment while determining unconscious decisions of IoE service aggregation. Subsequently, the virtual space of the XAI twin constructs a directed acyclic graph (DAG)-based Bayesian network that can infer a symbolic reasoning score over unconscious decisions through a first-order probabilistic language model. Furthermore, a Bayesian multi-arm bandits-based learning problem is proposed for reducing the gap between the expected explained score and current obtained score of the proposed neuro-symbolic XAI twin. To address the challenges of \emph{extensible}, \emph{modular}, and \emph{stateless management functions} in ZSM, the proposed neuro-symbolic XAI twin framework consists of two learning systems: 1) \emph{implicit learner} that acts as an unconscious learner in physical space, and 2) \emph{explicit leaner} that can exploit symbolic reasoning based on implicit learner decisions and prior evidence. Experimental results show that the proposed neuro-symbolic XAI twin can achieve around $96.26\%$ accuracy while guaranteeing from $18\%$ to $44\%$ more trust score in terms of reasoning and closed-loop automation.
\end{abstract}
	

	\IEEEpeerreviewmaketitle
	\section{INTRODUCTION}
	\label{INTRODUCTION}
	Explainable artificial intelligence (XAI) functions must be integrated into next-generation wireless networks such as 6G so as to enable an inherent analytical ability for characterizing the behavior of the Internet of Everything (IoE), inferring causes from the generated network data, and delivering correct processes to the right users \cite{IEEEhowto:D_Twin_1, IEEEhowto:Saad_6G_1, IEEEhowto:Ahmed_XAI_3, IEEEhowto:Guo_XAI_1, IEEEhowto:Tjoa_XAI_Survey_5, IEEEhowto:Khan_6G_2, IEEEhowto:Saad_Neuro_Sy}. Further, the next-generation network and service management functions of a wireless IoE have to fulfill the zero-touch network and service management (ZSM) by employing closed-loop intelligent automation of network and cross-domain service management for orchestration and control \cite{IEEEhowto:ETSI_Zero_Touch}. To enable zero-touch service data collection, data analytics, and intelligent automation of wireless IoE service aggregation, incorporating the capability of coping with geospatial features of each IoE service and understanding contextual metrics become essential \cite{IEEEhowto:XAI_Munir_ICC, IEEEhowto:Pant_IoE_3, IEEEhowto:Manogaran_IoE_2, IEEEhowto:Adhikari_IoE_4}. In this work, \emph{zero-touch} refers to a unique intelligent system for the wireless network that can autonomously capture the current networking environment and execute self-ruling communication and computational service decisions by interpreting the cause and effect of such decisions.

	\subsection{Motivation}
	Existing approaches for ZSM \cite{IEEEhowto:Chergui_Zero_6G_Slice, IEEEhowto:Grasso_AI_Zero, IEEEhowto:Chergui_AI_Zero, IEEEhowto:Rizwan_AI_Zero, IEEEhowto:Rezazadeh_A2C_Zero} are based on data-driven learning schemes. 
	Those are focused on hierarchical federated learning for network resource slicing \cite{IEEEhowto:Chergui_Zero_6G_Slice}, deep reinforcement learning (DRL) for computational resources management of flying ad-hoc networks \cite{IEEEhowto:Grasso_AI_Zero}, a statistical federated learning framework for network slicing \cite{IEEEhowto:Chergui_AI_Zero}, an unsupervised clustering for wireless network resource management and slicing \cite{IEEEhowto:Rizwan_AI_Zero}, and model-free DRL for virtual network function deployment \cite{IEEEhowto:Rezazadeh_A2C_Zero}. These \cite{IEEEhowto:Chergui_Zero_6G_Slice, IEEEhowto:Grasso_AI_Zero, IEEEhowto:Chergui_AI_Zero, IEEEhowto:Rizwan_AI_Zero, IEEEhowto:Rezazadeh_A2C_Zero} are not sufficient for enabling ZSM for wireless IoE because of various reasons.
	First, these prior approaches \cite{IEEEhowto:Chergui_Zero_6G_Slice, IEEEhowto:Grasso_AI_Zero, IEEEhowto:Chergui_AI_Zero, IEEEhowto:Rizwan_AI_Zero, IEEEhowto:Rezazadeh_A2C_Zero} are focused on a particular network and computational resource management function without consideration of service domain data collection, domain analytics based on collected data, and intelligence orchestration based on service domain analytics for wireless IoE. Second, the efforts in the prior art do not meet several fundamental principles of ZSM that are suggested by the European Telecommunications Standards Institute (ETSI) \cite{IEEEhowto:ETSI_Zero_Touch}, such as scalability, modularity, and stateless management functions on service provisioning. Third, the AI-based techniques in prior works do not comply with reliability and trustworthiness since they are unable to characterize the causes of strengths and weaknesses of their resource management decisions due to a lack of interpretability. 
	Therefore, to establish a ZSM for wireless IoE, there is a need for an intelligent system that can capture the current networking environment for decision making while also being  capable of \emph{reasoning causes}, \emph{analyzing effects} \cite{IEEEhowto:Saad_Neuro_Sy}, and executing autonomous decisions with a \emph{higher reliability}.
	\subsection{Goal and Challenges}
	The main goal of this work is to develop an \emph{explainable artificial intelligence twin} system for the network service provider that assures the reasoning and trustworthiness of zero-touch IoE service management. To guarantee the reasoning and trustworthy operation of zero-touch IoE, an AI system must maintain the observing capabilities of physical behavior through the neural-networks-driven models and preserve the interpretability by a rule-based symbolic reasoning scheme. 
	In other words, we will focus on a mechanism that can capture and observe the time-dependent physical behavior of wireless IoE from the current networking dynamics, service users' requirements, and contextual metrics.
	Then the XAI system will be capable of correcting erroneous decisions on service user association, uplink, downlink data rates control, and service provisioning based on its reasoning. 
	In order to do this, we must address several challenges. First, characterizing correlation among the networking dynamics \cite{IEEEhowto:XAI_Munir_ICC, IEEEhowto:Pant_IoE_3, IEEEhowto:Manogaran_IoE_2, IEEEhowto:Adhikari_IoE_4} such as reference signal received power (RSRP), user uplink and downlink data rates, communication and computational capacity, user speed, signal-to-interference-plus-noise ratio (SINR), and reference signal received quality (RSRQ) is a key challenge due to distinct communication and computational requirements of each IoE service. Second, determining the most contributing coefficient of network parameters for zero-touch IoE service aggregation decisions is another challenge since service data, user mobility, service processes, and the capacity of physical devices are not fixed and are random over time. Last but not least, in the traditional AI mechanisms, it is hard to balance the accuracy and interoperability tradeoff of zero-touch IoE service management for deciding service user association, uplink, downlink data rates control, and service provisioning. 
	
	\subsection{Summary of contributions}
	The main contribution of this paper is, therefore, a novel neuro-symbolic explainable artificial intelligence twin framework for zero-touch IoE network and service management. The proposed approach creates an AI system that can reason the root causes of each IoE service requirement and execution decision, analyze the effects of reasoning, achieve a certain level of channel quality index (CQI), and provide autonomous IoE service association and data rates control to IoE service. Our main contributions include:
	\begin{itemize}
		\item We introduce a novel XAI-twin framework for ZSM in wireless networks that can characterize both the reasoning and effects for achieving higher accuracy, data efficiency, transparency, and trustworthy IoE service execution decisions. Thus, the physical space of the XAI twin can capture the dynamic environment of network parameters and contextual metrics of each IoE element to devise an unconscious decision on communication and computation resources, meanwhile, each of the decisions is observed evidently by a virtual space to secure a closed-loop IoE service execution with reliability. 
		
		\item For capturing the current IoE environmental dynamics of each physical space, we propose a multivariate regression approach by developing a neural-networks-driven model to determine unconscious decisions for IoE service association, uplink, and downlink data rates control since these outcomes rely entirely on time-variant network parameters. The unconscious decision can not do introspection due to its trained memory.
		Then, we construct a directed acyclic graph-based Bayesian network on each network parameter and contextual feature that can infer a symbolic reasoning score of physical space decisions through a first-order probabilistic language model. In particular, the symbolic reasoning score is determined by the marginalized joint probability distribution of all network parameters and contextual metrics.
		
		\item Then, we formulate a decision problem, where the main objective is to minimize the residual between an expected explained score and the current obtained score on the corresponding decisions in virtual space by coordinating with each physical space. The formulated problem is reduced to a Bayesian multi-arm bandits \cite{IEEEhowto:MAB_Book_1}, in which, each physical space of XAI twin is considered as a bandit arm given that its unconscious decisions and probabilistic evidence.      
		
		\item Next, we develop a neuro-symbolic XAI twin framework that consists of two learning systems: 1) \emph{implicit learner} that acts as an unconscious decision-maker in physical space via a neural-networks-driven model and 2) \emph{explicit leaner} that exploits symbolic reasoning based on implicit learner decisions and prior knowledge. In particular, the explicit learner finds reasoning of the network dynamics and contextual metrics and executes a Bayesian multi-arm bandits system for accomplishing the corrected IoE service execution decisions based on the reasoning and reliability.  
		
		\item Through experiments using a state-of-the-art $5$G dataset (B$\_2020.02.13\_13.03.24$) \cite{IEEEhowto:Data_set_5g}, we show that our neuro-symbolic XAI twin framework can outperform the other benchmarking approaches in terms of accuracy, explainable score, and closed-loop IoE service execution. In particular, the developed neuro-symbolic XAI twin framework affords around $96.26\%$ accuracy for both uplink and downlink rates allocation while yielding $44\%$ and $18\%$ more reasoning score than that of the Gradient-based bandits and Epsilon-greedy based schemes for closed-loop IoE service execution. Additionally, the proposed approach can provide higher CQI to IoE services while $72\%$ of IoE service can achieve a CQI over $10$.
	\end{itemize}
	
	The rest of this paper is organized as follows. In Section \ref{Prior_Works}, we present a brief literature review based on the prior research. We present the proposed system model of neuro-symbolic explainable artificial intelligence twin for zero-touch network and service management in Section \ref{System_Model} and important notations are summarized in Table \ref{sum_not}. In Section \ref{Learning_Problem_XAI_Twin}, we formalize the learning problem of XAI twin. Then, our neuro-symbolic XAI twin framework is designed in Section \ref{NS_XAI_Twin_Framework}. Experimental results are analyzed in Section \ref{Experimental_Result}, and conclusions are drawn in Section \ref{Conclusion}. 
	
	\section{Prior Works}
	\label{Prior_Works}
	In this section, we provide a brief discussion on some of the interesting prior works and their shortcomings that are addressed in the proposed neuro-symbolic explainable artificial intelligence twin framework for zero-touch IoE network and service management.
	
	\subsection{{Role of AI Towards ZSM}}
	Recently, several works \cite{IEEEhowto:Chergui_Zero_6G_Slice, IEEEhowto:Grasso_AI_Zero, IEEEhowto:Chergui_AI_Zero, IEEEhowto:Rizwan_AI_Zero, IEEEhowto:Rezazadeh_A2C_Zero} investigated the role of AI for zero-touch network and service management in beyond 5G wireless networks. In \cite{IEEEhowto:Chergui_Zero_6G_Slice}, the authors proposed a distributed management and orchestration framework using a hierarchical AI-driven closed-loop control system for addressing the challenge of network slicing in large-scale networks. However, this prior work does not capture the network dynamics such as IoE service data heterogeneity, CQI, and service user mobility. The work in \cite{IEEEhowto:Grasso_AI_Zero} developed a DRL-based zero-touch adaptation approach for computing task offloading in a drone system. However, this approach is not scalable because this only works for the predefined fixed number of drone deployments.
	The authors in \cite{IEEEhowto:Chergui_AI_Zero} proposed a statistical federated learning approach for closed-loop network resource slicing automation on different technological domains. However, the technical approach was limited mostly to qualitative discussions.
	The work in \cite{IEEEhowto:Rizwan_AI_Zero} used a k-means clustering approach integrated with domain expert feedback to perform network management. However, this work does not take into account network service association and data rates control.
	The authors in \cite{IEEEhowto:Rezazadeh_A2C_Zero} investigated a continuous model-free DRL approach for minimizing the system's energy consumption and virtual network function deployment cost. This approach is not appropriate when extreme reasoning becomes more important for enabling a reliable automated network and service management and meeting  the key performance indicator (KPI) of ZSM. 
	In summary, the works \cite{IEEEhowto:Chergui_Zero_6G_Slice, IEEEhowto:Grasso_AI_Zero, IEEEhowto:Chergui_AI_Zero, IEEEhowto:Rizwan_AI_Zero, IEEEhowto:Rezazadeh_A2C_Zero} have attempted to define a couple of principles towards ZSM such as closed-loop automation and scalability; however, they did  not consider \emph{extensibility}, \emph{modularity}, and \emph{stateless management functions} \cite{IEEEhowto:ETSI_Zero_Touch} through  reasoning.

	\subsection{{IoE Service Management}}
	Recently, some of the challenges related to IoE network and service management over wireless networks were studied in \cite{IEEEhowto:XAI_Munir_ICC, IEEEhowto:Pant_IoE_3, IEEEhowto:Manogaran_IoE_2, IEEEhowto:Adhikari_IoE_4}. The work in \cite{IEEEhowto:XAI_Munir_ICC} proposed an XAI-enabled IoE service delivery framework by designing a multivariant regression problem. The proposed framework can enable both intelligence and interpretation during the IoE service delivery decisions. In \cite{IEEEhowto:Pant_IoE_3}, the authors investigated the problem of malicious and benign nodes detection for an IoE network by proposing a machine learning architecture. 
	The authors in \cite{IEEEhowto:Manogaran_IoE_2} proposed a mutable service distribution model based on deep recurrent learning to enable unified IoE service response by preventing service overlapping. In \cite{IEEEhowto:Adhikari_IoE_4}, the authors used DRL to design a cybertwin-enabled edge framework for finding a strategy based on dynamic IoE service requirements.
	However, these works \cite{IEEEhowto:XAI_Munir_ICC, IEEEhowto:Pant_IoE_3, IEEEhowto:Manogaran_IoE_2, IEEEhowto:Adhikari_IoE_4} do not investigate the root causes and effects on wireless IoE service association and data rates assignment, nor do they meet \emph{self-adaptive} service requirements.
	
	\subsection{{Digital Twin and Learning}}
	The use of learning in digital twins (DTs) was studied in \cite{IEEEhowto:Xu_Twin_2, IEEEhowto:Prathiba_Twin_3,IEEEhowto:Zhu_Twin_6, IEEEhowto:Walid_Twin_5}
	In \cite{IEEEhowto:Xu_Twin_2}, a cybertwin-assisted asynchronous federated learning (AFL) scheme was proposed for communication and computation resource allocation in an edge computing network, where the role of a cybertwin is to coordinate between each client learning model and cloud server for aggregation during the training process. 
	The work in \cite{IEEEhowto:Prathiba_Twin_3} developed a new reinforcement learning approach for a personalized vehicular service provision scheme in 6G vehicle-to-everything (6G-V2X). 
	The work in \cite{IEEEhowto:Zhu_Twin_6} investigated a digital twin framework by proposing a federated Markov chain Monte Carlo model to solve the challenge of distributed data privacy for federated analytics. In \cite{IEEEhowto:Walid_Twin_5}, the authors developed a continual learning approach to enable the operation of DTs in non-stationary environments.
	While the studies in \cite{IEEEhowto:Xu_Twin_2, IEEEhowto:Prathiba_Twin_3, IEEEhowto:Zhu_Twin_6, IEEEhowto:Walid_Twin_5} have incorporated the concept of DT for solving interesting research challenges from an edge learning perspective in stationary and no-stationary settings, they have not investigated the role of DT in analyzing the \emph{interoperability} of the AI models' decisions for a wireless network. 
	Therefore, this work develops a novel neuro-symbolic XAI twin for wireless IoE. The physical space of the XAI twin executes a neural network-driven multivariate regression to capture the time-dependent wireless IoE environment, and the virtual space of the XAI twin constructs a directed acyclic graph (DAG)-based Bayesian network to infer a symbolic reasoning score over physical space decisions. The system model of XAI twin for wireless network is presented in the following section.

	\section{System Model of XAI Twin for Wireless Network}
	\label{System_Model}
	\begin{figure}[!t]
		\centerline{\includegraphics[scale=.6]{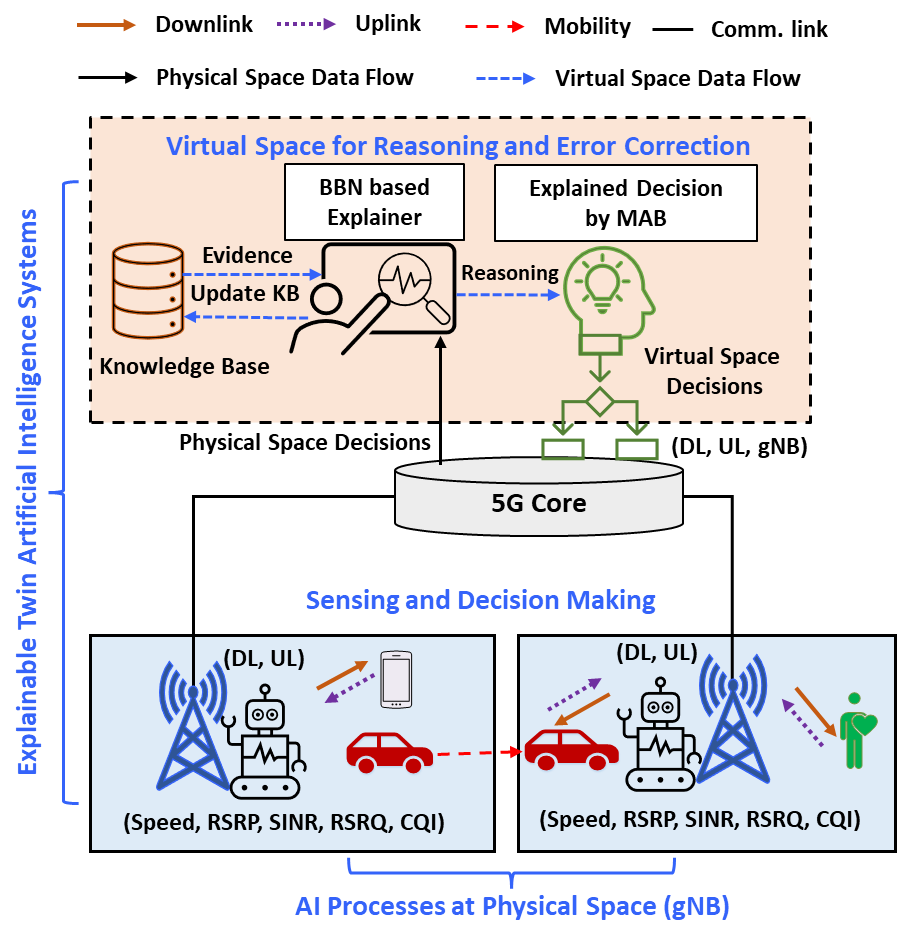}}
		\caption{A system model of neuro-symbolic explainable artificial intelligence twin for zero-touch network and service management.}
		\label{System_model}
	\end{figure}
	\subsection{System Overview}
	\label{System_Overview}
	We consider a wireless network having a set $\mathcal{G}$ of $G$  gNBs are physically deployed to serve the network users in a certain geographical area. Each gNB $g \in \mathcal{G}$ is equipped with a computational server to enable a set $\mathcal{I}$ of $I$ IoE services. Each gNB $g \in \mathcal{G}$ is connected to a core network (i.e., $5$G core) through wired connectivity. In order to enable connected intelligence in this system, we consider a \emph{neuro-symbolic XAI twin} of the entire network as seen in Figure \ref{System_model}. In particular, each gNB $g \in \mathcal{G}$ is considered as a physical space of XAI twin that can capture current time-dependent networks parameters of network entities and execute a neural-network driven multivariate regression for solving networking problems, such as IoE service association, uplink, and downlink data rates control. The virtual side of the XAI twin in the $5$G core can perform reasoning and error correction on physical space decisions based on prior knowledge by enabling symbolic reasoning.
	
	The network's dynamics depend on both states of wireless node (i.e., gNB) and service user equipment. In particular, network parameters such as user mobility (i.e., speed) $s$, RSRP $p$, RSRQ $q$, SINR $\delta$, and CQI $c$ are prominent contextual features for deciding on the downlink data rate, uplink data rate, and service user association to the gNB. 
	Each IoE service $i \in \mathcal{I}$ at gNB $g \in \mathcal{G}$ can serve a set $\mathcal{J}_i$ of $J_i$ user. Therefore, the network and service dynamics of each user $j \in \mathcal{J}_i$ will have distinct requirements such as downlink data transfer rate $d_{ij}^\textrm{req}$, uplink data transfer rate $u_{ij}^\textrm{req}$, communication and computational delay  $\tau_{ij}^\textrm{req}$, and a tuple $X \coloneqq (s_{ij}^\textrm{phy}, p_{ij}^\textrm{phy}, q_{ij}^\textrm{phy}, \delta_{ij}^\textrm{phy}, c_{ij}^\textrm{phy})$ of $N$ contextual features in the physical space. Here, $s_{ij}^\textrm{phy}$, $p_{ij}^\textrm{phy}$, $q_{ij}^\textrm{phy}$, $\delta_{ij}^\textrm{phy}$, and $c_{ij}^\textrm{phy}$ represent speed, RSRP, RSRQ, SINR, and CQI, respectively of an IoE  service user $j \in \mathcal{J}_i$. Each IoE service user $j \in \mathcal{J}_i$ sends $\alpha_{ij}^\textrm{req}$ [Mb] data to gNB $g \in \mathcal{G}$ via uplink communication while the computational unit of gNB $g \in \mathcal{G}$ processes that request based on service $i \in \mathcal{I}$. 
	Subsequently, user $j \in \mathcal{J}_i$ receives $\beta_{ij}^\textrm{req}$ [Mb] through the downlink communication from gNB $g \in \mathcal{G}$. 
	In the above scenario, the challenge is to allocate the downlink data transfer rate $d_{ij}^\textrm{req}$ and uplink data transfer rate $u_{ij}^\textrm{req}$ to the user $j \in \mathcal{J}_i$ when maintaining a certain level of CQI $c_{ij}^\textrm{phy}$ under the service completion delay requirement $\tau_{ij}^\textrm{req}$ due to service heterogeneity.
	Additionally, user mobility (i.e., speed) $s_{ij}^\textrm{phy}$ also affects the CQI since the service user association will be shifted from one gNB $g \in \mathcal{G}$ to another gNB $g+1 \in \mathcal{G}$ due to its location changes.
	
	\begin{table}[t!]
		\caption{Summary of Notations }
		\begin{center}
			\begin{tabular}{|p{1.5cm}|p{6cm}|}
				\hline
				\textbf{Notation}&{\textbf{Description}} \\
				\hline
				$\mathcal{G}$ & Set of next generation NodeBs (gNBs) \\
				\hline
				$\mathcal{I}$ & Set of IoE services\\
				\hline
				$\mathcal{J}_i$ & Set of users for IoE service $i \in \mathcal{I}$ \\
				\hline
				$d_{ij}^\textrm{req}$ [Mbps] & Required downlink data transfer rate of user 
				$j \in \mathcal{J}_i$  \\
				\hline
				$u_{ij}^\textrm{req}$ [Mbps] & Required uplink data transfer rate of user $j \in \mathcal{J}_i$\\
				\hline
				$\tau_{ij}^\textrm{req}$ & Communication and computational tolerable delay of user $j \in \mathcal{J}_i$  \\
				\hline
				$X$ & A tuple for contextual features on physical space of user $j \in \mathcal{J}_i$\\
				\hline
				$s_{ij}^\textrm{phy}$ [km/h] & User mobility speed of user $j \in \mathcal{J}_i$\\
				\hline
				$p_{ij}^\textrm{phy}$ [dBm] & Reference signal received power of user $j \in \mathcal{J}_i$ \\
				\hline
				$q_{ij}^\textrm{phy}$ [dB]& Reference signal received quality of user $j \in \mathcal{J}_i$ \\
				\hline
				$\delta_{ij}^\textrm{phy}$ [dB] & Signal-to-interference-plus-noise ratio of user $j \in \mathcal{J}_i$\\
				\hline
				$c_{ij}^\textrm{phy}$& Channel quality indicator\\
				\hline
				$\alpha_{ij}^\textrm{req}$ [Mb] & Amount of uploaded data for user $j \in \mathcal{J}_i$ \\
				\hline
				$\beta_{ij}^\textrm{req}$ [Mb]& Amount of downloaded data for user $j \in \mathcal{J}_i$\\
				\hline
				$u_{ij}^\textrm{exe}$ [Mbps] & Allocated uplink data rate for user $j \in \mathcal{J}_i$\\
				\hline
				$d_{ij}^\textrm{exe}$ [Mbps] & Allocated downlink data rate for user $j \in \mathcal{J}_i$\\
				\hline
				$\boldsymbol{X}$ & Context matrix with a dimension of $J \times (N+1)$ at the physical space of XAI twin\\
				\hline
				$\boldsymbol{\xi}$ & Parameters matrix with a dimension of $(N+1) \times 2$ at the physical space of XAI twin\\
				\hline
				$\boldsymbol{Y}$ & $J\times2$ dimensional output matrix at the physical space of XAI twin\\
				\hline
				$\boldsymbol{\epsilon}$ &  $J\times2$ matrix with zero mean and covariance $\Xi$ at the physical space of XAI twin. \\
				\hline
				$T^\textrm{up}_{ij}$ & Uplink data transfer duration for user $j \in \mathcal{J}_i$ of IoE service $i$\\
				\hline
				$T^\textrm{queue}_{ij}$ & Queuing waiting duration of user $j \in \mathcal{J}_i$ for IoE service execution $i$ \\
				\hline
				$T^\textrm{exe}_{ij}$ & Execution duration for user $j \in \mathcal{J}_i$ of IoE service $i$ \\
				\hline
				$T^\textrm{down}_{ij}$ & Downlink data transfer duration for user $j \in \mathcal{J}_i$ of IoE service $i$\\
				\hline
				$\tau_{ij}^\textrm{tot}$ & The overall delay of of IoE service $i$ fulfillment for user $j \in \mathcal{J}_i$   \\
				\hline
				$\gamma_i$ & Computational capacity for IoE service $i$ at gNB $g \in \mathcal{G}$ \\
				\hline
				$\lambda_i$ & Arrival rate of IoE service request $i$ at gNB $g \in \mathcal{G}$   \\
				\hline
				$\mu_i$ & Computational rate for IoE service $i$ at gNB $g \in \mathcal{G}$  \\
				\hline
				$\omega_i$ & Server utilization rate for IoE service $i$ at gNB $g \in \mathcal{G}$\\
				\hline
				$W$ & A tuple of random variables for $M$ network dynamics and contextual parameters\\
				\hline
				$\boldsymbol{E}$ &Evidence of all observed network dynamics and contextual metric parameters at virtual space of XAI twin\\
				\hline
				$\boldsymbol{Z}$ & Unobserved evidence of network dynamics and contextual metric parameters at virtual space of XAI twin\\
				\hline
				$\boldsymbol{\theta}$ & knowledge base (KB) at virtual space of XAI twin\\
				\hline
				$\phi$ & Exploration coefficient for learning at virtual space of XAI twin\\
				\hline
				$\chi^\textrm{up}$ [MHz]& Bandwidth for uplink communication \\
				\hline
				$\chi^\textrm{down}$ [MHz]& Bandwidth for downlink communication \\
				\hline
				$\eta$& Minimum level of channel quality indicator\\
				\hline
				$\omega^\textrm{max}$& Maximum computational capacity for executing IoE service $i$ at $g \in \mathcal{G}$\\
				\hline
				$a_{ij}^g$& gNB selection decision indicator \\
				\hline
			\end{tabular}
			\label{sum_not}
		\end{center}
		\vspace{-4mm}
	\end{table}

	\subsection{Network and Service Dynamics in Physical Space of XAI Twin}
	\label{PhysicalSpaceofXAITwin}
	Each gNB $g \in \mathcal{G}$ is considered as a physical space of XAI twin. Here, our goal is to capture the contextual features of the network and service dynamics for allocating uplink $u_{ij}^\textrm{exe}$ and downlink $d_{ij}^\textrm{exe}$ data rates while maintain a certain level of CQI. 
	Therefore, we consider a tuple $X \coloneqq (s_{ij}^\textrm{phy}, p_{ij}^\textrm{phy}, q_{ij}^\textrm{phy}, \delta_{ij}^\textrm{phy}, c_{ij}^\textrm{phy})$ of $N$ contextual features that are already known in physical space, where the procedure for calculating the contextual features $X$ follows the $3$GPP standard \cite{IEEEhowto:3GPP_1, IEEEhowto:5G_Architecture_3PPP}. 
	Then, we can formalize a multivariate regression \cite{IEEEhowto:Multivariate_Regression_1, IEEEhowto:Multivariate_Regression_2, IEEEhowto:Multivariate_Regression_3} for capturing the coefficients of $N$ contextual features, and thus, we can get a $J\times2$ response matrix $\boldsymbol{Y}$ for allocating uplink $u_{ij}^\textrm{exe}$ and downlink $d_{ij}^\textrm{exe}$ data rates to the IoE service users.
	The system of equations of a multivariate regression \cite{IEEEhowto:Multivariate_Regression_1, IEEEhowto:Multivariate_Regression_2, IEEEhowto:Multivariate_Regression_3} in the physical space of XAI twin can be defined as follows:     
	\begin{equation} \label{eq:Mul_Var_Reg}
		\begin{split}
			\underbrace{\begin{pmatrix}
					u_{i1}^\textrm{exe} & d_{i1}^\textrm{exe} \\
					u_{i2}^\textrm{exe} & d_{i2}^\textrm{exe} \\
					\vdots & \vdots \\
					u_{iJ}^\textrm{exe} & d_{iJ}^\textrm{exe} 
			\end{pmatrix}}_{\boldsymbol{Y}}
			=
			\underbrace{\begin{pmatrix}
					1 & s_{i1}^\textrm{phy} &  p_{i1}^\textrm{phy} &  q_{i1}^\textrm{phy} &  \delta_{i1}^\textrm{phy} &  c_{i1}^\textrm{phy} \\
					1 & s_{i2}^\textrm{phy} &  p_{i2}^\textrm{phy} &  q_{i2}^\textrm{phy} &  \delta_{i2}^\textrm{phy} &  c_{i2}^\textrm{phy} \\
					\vdots & \vdots &  \vdots &  \vdots & \vdots &  \vdots \\
					1 & s_{iJ}^\textrm{phy} &  p_{iJ}^\textrm{phy} &  q_{iJ}^\textrm{phy} &  \delta_{iJ}^\textrm{phy} &  c_{iJ}^\textrm{phy}
			\end{pmatrix}}_{\boldsymbol{X}} \\ \\
			\underbrace{\begin{pmatrix}
					\xi_{u_{i0}^\textrm{exe}} & \xi_{d_{i0}^\textrm{exe}} \\
					\xi_{u_{i1}^\textrm{exe}} & \xi_{d_{i1}^\textrm{exe}} \\
					\vdots & \vdots \\
					\xi_{u_{iN}^\textrm{exe}} & \xi_{d_{iN}^\textrm{exe}} \\
			\end{pmatrix}}_{\boldsymbol{\xi}}
			+
			\underbrace{\begin{pmatrix}
					\epsilon_{u_{i1}^\textrm{exe}} & \epsilon_{d_{i1}^\textrm{exe}} \\
					\epsilon_{u_{i2}^\textrm{exe}} & \epsilon_{d_{i2}^\textrm{exe}} \\
					\vdots & \vdots \\
					\epsilon_{u_{iJ}^\textrm{exe}} & \epsilon_{d_{iJ}^\textrm{exe}} 
			\end{pmatrix}}_{\boldsymbol{\epsilon}},
		\end{split}
	\end{equation}
	where, $\boldsymbol{X}$ is a $J \times (N+1)$ matrix of the contextual features $X$, $\boldsymbol{\xi}$ is a parameters matrix with a dimension of $(N+1) \times 2$, and $\boldsymbol{\epsilon}$ is a $J\times2$ matrix with zero mean and covariance $\boldsymbol{\Omega}$. By solving the system of equations in \eqref{eq:Mul_Var_Reg}, we can allocate uplink $u_{ij}^\textrm{exe}$ and downlink $d_{ij}^\textrm{exe}$ data rates for each IoE service user $j \in \mathcal{J}_i$. 
	
	We define $\tau_{ij}^\textrm{req}$ as an end-to-end (E2E) tolerable delay requirement for IoE service user $j \in \mathcal{J}_i$. Thus, to assure E2E maximum tolerable delay of each IoE service execution, it is essential to characterize four types of delays such as uplink data transfer duration $T^\textrm{up}_{ij}$, waiting time duration $T^\textrm{queue}_{ij}$ for computation, computational duration $T^\textrm{exe}_{ij}$, and downlink data transfer duration $T^\textrm{down}_{ij}$. For an uplink data rate $u_{ij}^\textrm{exe}$, each IoE service user sends an amount of data $\alpha_{ij}^\textrm{req}$ to gNB $g\in \mathcal{G}$ and the uplink data transfer duration is $\tau_{ij}^{\textrm{up}} =\frac{\alpha_{ij}^\textrm{req}}{u_{ij}^\textrm{exe}} $. The arrival of each amount of data $\alpha_{ij}^\textrm{req}$ at gNB $g\in \mathcal{G}$ follows a Poisson distribution with a probability $P(\tau_{ij}^{\textrm{up}})$. Therefore, the mean uplink duration will be $\mathbb{E}[T^\textrm{up}_{ij}]= \sum_{\forall j \in \mathcal{J}_i} \tau_{ij}^{\textrm{up}} P(\tau_{ij}^{\textrm{up}})$ at gNB $g\in \mathcal{G}$.
	
	We consider that each IoE service $i \in \mathcal{I}$ is computed at a dedicated computational host at gNB $g \in \mathcal{G}$ with a computational capacity $\gamma_i$. Each IoE user task $j \in \mathcal{J}_i$ arrives at gNB $g\in \mathcal{G}$ by following a Poisson process \cite{IEEEhowto:Darabi_MG1} with an arrival rate of $\lambda_i$. Then, we can find the service computational rate $\mu_i = \frac{\lambda_i \alpha_{ij}^\textrm{req}}{\gamma_i}$, and the service time $T_{ij}^\textrm{service}= \frac{1}{\mu_i}$ of IoE user $j \in \mathcal{J}_i$ will follow a general holding time distribution. For an arrival rate of $\lambda_i$ and service computational rate $\mu_i$, the server utilization rate becomes $\omega_i = \sum_{\forall j \in \mathcal{J}_i} \frac{\lambda_i}{\mu_i}$. In this computational delay model, we consider an \emph{M/G/1} queuing system \cite{IEEEhowto:MG1_1, IEEEhowto:Darabi_MG1} which is suitable here since the arrival of IoE task follows a Poisson process and service time of the computational host at gNB $g \in \mathcal{G}$ belongs to a general distribution. By applying the Pollaczek-Khinchin mean formula \cite{IEEEhowto:Pollaczek_Khinchine_Formula, IEEEhowto:Munir_APNOMS_Int_Ser}
	the expectation of waiting time of IoE service can be calculated as $\mathbb{E}[T^\textrm{queue}_{ij}] = \frac{\lambda_i \mathbb{E}[(T_{ij}^\textrm{service})^2]}{2(1-\omega_i)}$, and thus, computational delay at gNB $g \in \mathcal{G}$ becomes $\mathbb{E}[T^\textrm{exe}_{ij}] = \mathbb{E}[T_{ij}^\textrm{service}] + \frac{\lambda_i \mathbb{E}[(T_{ij}^\textrm{service})^2]}{2(1-\omega_i)}$. For squared coefficient of variation $C^2 = \frac{\mathbb{V}[T_{ij}^\textrm{service}]}{\mathbb{E}[T_{ij}^\textrm{service}]^2}$, the total delay for computation is calculated as $\mathbb{E}[T^\textrm{exe}_{ij}] = (1+ \frac{1+C^2}{2} \frac{\omega_i}{1-\omega_i})\mathbb{E}[T_{ij}^\textrm{service}]$. Here, to fulfill each IoE service execution, the consider system requires to send $\beta_{ij}^\textrm{req}$ amount of data to the IoE service user $j \in \mathcal{J}_i$. For a downlink data rate $d_{ij}^\textrm{exe}$, the duration of transferring $\beta_{ij}^\textrm{req}$ can be calculated as $\tau_{ij}^{\textrm{down}} =\frac{\beta_{ij}^\textrm{req}}{d_{ij}^\textrm{exe}}$. Then, the mean downlink time will be $\mathbb{E}[T^\textrm{down}_{ij}]= \sum_{\forall j \in \mathcal{J}_i} \tau_{ij}^{\textrm{down}} P(\tau_{ij}^{\textrm{down}})$, where $P(\tau_{ij}^{\textrm{down}}$ is a probability of departure duration $\tau_{ij}^{\textrm{down}}$ and that follows a Poisson distribution.   
	Thus, an average E2E duration for executing each IoE service by gNB $g \in \mathcal{G}$ is determined as follows:     
	\begin{equation} \label{eq:j5_tot_exe_delay}
		\begin{split}
			\mathbb{E}[\tau_{ij}^\textrm{tot}] = \mathbb{E}[T^\textrm{up}_{ij}] + (1+ \frac{1+C^2}{2} \frac{\omega_i}{1-\omega_i})\mathbb{E}[T_{ij}^\textrm{service}] + \mathbb{E}[T^\textrm{down}_{ij}],
		\end{split}
	\end{equation}  
	where $\mathbb{E}[\tau_{ij}^\textrm{tot}]$ must be smaller than or equal to the  communication and computational delay requirement $\tau_{ij}^\textrm{req}$, $\mathbb{E}[\tau_{ij}^\textrm{tot}] \le \tau_{ij}^\textrm{req}$. An user $j \in \mathcal{J}_i$ of IoE service $i \in \mathcal{I}$ does not know which gNB $g \in \mathcal{G}$ becomes suitable for desired service execution while nearby gNB can receive service request simultaneously. Additionally, the physical space environment of each at each $g \in \mathcal{G}$ is independent and distinct. Therefore, to allocate uplink $u_{ij}^\textrm{exe}$ and downlink $d_{ij}^\textrm{exe}$ data rates along with appropriate gNB $g \in \mathcal{G}$, it is imperative to characterize the causes and effects of the current network and service dynamics. 
	For assuring sufficient uplink and downlink data rates to IoE service, the reasoning for the effects of user mobility, RSRP, SINR, and RSRQ on CQI become essential.
	Therefore, the reasoning of the IoE service execution decisions can be determined by utilizing the concept of neuro-symbolic AI. The detailed description of the reasoning mechanisms are formalized in the next subsection.
	
	\subsection{Reasoning in Virtual Space of XAI Twin}
	\label{VirtualSpaceofXAITwin}
	The role of the virtual space within the XAI twin is to reason the root causes and accomplish an updated allocation decision for the considered IoE service execution. 
	We consider a Bayesian network (BN) \cite{IEEEhowto:Bayesian_Log_Pro} with first-order probabilistic languages \cite{IEEEhowto:FOPL_Base} to capture a logical reasoning by encoding influences among the contextual metrics and network dynamics. Thus, in the virtual space, a directed acyclic graph (DAG) \cite{IEEEhowto:Bayesian_Log_Pro} is considered for characterizing quantitative probability densities of network dynamics and contextual parameters such as user mobility, RSRP, SINR, RSRQ, and CQI.
	
	We define $M$ network dynamics and contextual parameters as a tuple of random variables $W = (W_1, W_2, \dots, W_M)$ and each random variable $W_m$ is indexed by $m$, where $m \in (1, \dots, M)$. Each random variable $W_m$ follows bimodal distribution due to the distinct requirements and physical condition of IoE services. The random variables $W$ are physically known metrics and they are observed by physical space. Thus, the tuple of random variables $W$ includes user speed $s_{ij}^\textrm{phy}$, RSRP $p_{ij}^\textrm{phy}$, RSRQ $q_{ij}^\textrm{phy}$, SINR $\delta_{ij}^\textrm{phy}$, CQI $c_{ij}^\textrm{phy}$, uplink data rate $u_{ij}^\textrm{exe}$, downlink data rate $d_{ij}^\textrm{exe}$, and user association of gNB $a_{ij}^g$ for all users in the considered network. To build a BN, we consider a directed DAG $D$ that is encompassed with $M$ nodes, where each node $W_m$ denotes a random variable of each network dynamics and contextual parameter. Therefore, each directional edge will head from a cause contextual metric node (i.e., random variable) to an effect contextual metric node. Thus, the parents $\boldsymbol{Pa}(W_m)$ of each parameter node $W_m$ will be all nodes that are pointing towards node $W_m$. The factorization of a joint probability distribution (JPD) of the network dynamics and contextual parameters is as follows \cite{IEEEhowto:Bayesian_Log_Pro}: 
	\begin{equation} \label{eq:prob_context_matric}
		\begin{split}
			P(W_1, W_2, \dots, W_M) = P(W_1)P(W_2|W_1, W_3) \\P(W_3|W_1, W_2) \dots P(W_M|W_1, W_{M-1}).
		\end{split}
	\end{equation} 
	The JPD of the network dynamics and contextual parameters \eqref{eq:prob_context_matric} contains both descendants and non-descendants. However, our goal is to characterize the JPD with respect to descendants of each parameters. Therefore, by removing non-descendants of $W_m$, we can rewrite \eqref{eq:prob_context_matric} as, 
	\begin{equation} \label{eq:descendants}
		\begin{split}
			P(W_1, W_2, \dots, W_M) = \\P(W_1|\boldsymbol{Pa}(W_n)) \dots P(W_M|\boldsymbol{Pa}(W_M)).
		\end{split}
	\end{equation} 
	\eqref{eq:descendants} represents a generic model of JPD for the considered network dynamics and contextual parameters in the virtual space of the XAI twin. Therefore, the network structure completely follows a DAG since \eqref{eq:descendants} is a factorization of the JPD of all network dynamics and contextual metrics parameters $W = (W_1, W_2, \dots, W_M)$. In particular, for random variables $W = (W_1, W_2, \dots, W_M)$, the JPD can be written as follows:
	\begin{equation} \label{eq:JPD}
		\begin{split}
			P(W_1, W_2, \dots, W_M) = \prod_{m=1}^{M} P(W_m|\boldsymbol{Pa}(W_m)),
		\end{split}
	\end{equation}  
	where $\boldsymbol{Pa}(W_m)$ represents a set of parents for each parameter node $W_m$.
	
	In this model, we consider both conditional probability queries (CPQs) and maximum a posterior probability (MAP) queries for inference since CPQ can handle many interconnected variables from distribution while MAP can estimate the mode of a posterior distribution. 
	We define $\boldsymbol{E}$ as the vector of evidence values of all observed network dynamics and contextual metric parameters, where each evidence $\boldsymbol{e} \in \boldsymbol{E}$ can query on parameter $W_m$. As a result, unobserved (i.e., non-evidence) network dynamics and contextual metric parameters $\forall \boldsymbol{z} \in  \boldsymbol{Z}$ can be calculated by utilizing the known evidences $\boldsymbol{E}$. Therefore, we can find the reason of unobserved network dynamics and contextual metric variables $\forall \boldsymbol{z} \in \boldsymbol{Z}$, as follows \cite{IEEEhowto:BN_Base}: 
	\begin{equation} \label{eq:reasoning}
		\begin{split}
			P(W_m|\boldsymbol{e}) = \frac{P(W_m,\boldsymbol{e})}{P(\boldsymbol{e})} \propto \sum_{\boldsymbol{z} \in \boldsymbol{Z}} P(W_m, \boldsymbol{e}, \boldsymbol{z}).
		\end{split}
	\end{equation} 
	In other words, we marginalize the JPD over the unobserved network dynamics and contextual metric variables $\boldsymbol{Z} \in W \setminus \boldsymbol{E}$ for inference. Then, we determine most probable explanation $\Lambda$ for $\boldsymbol{Z}$ of all network dynamics and contextual metric variables $W$ except observed $\boldsymbol{E}$ as follows \cite{IEEEhowto:BN_Base}: 
	\begin{equation} \label{eq:most_probable_explanation}
		\begin{split}
			P(\Lambda) = \underset{\forall \boldsymbol{z} \in \boldsymbol{Z}}{\argmax} \; P_{\boldsymbol{e} \sim \boldsymbol{E}}(\boldsymbol{z}|\boldsymbol{e}),
		\end{split}
	\end{equation} 
	where \eqref{eq:most_probable_explanation} can determine a statistical explanation for the root cause of candidate uplink $u_{ij}^\textrm{exe}$ and downlink $d_{ij}^\textrm{exe}$ data rates for each IoE service uses. Additionally, the virtual space of the XAI twin generates the corresponding candidate gNBs $\boldsymbol{a}, \forall a_{ij}^g \in \boldsymbol{a}$ towards IoE service association. Thus, we can construct a knowledge base (KB) $\boldsymbol{\theta}$ using the proposed BBN-based explainer for the virtual space of XAI twin, where for given $W$, KB of each user represents by $\theta_{i,j}^{g}$ and $\forall \theta_{i,j}^{g} \in \boldsymbol{\theta}$. Now, based on the reasoning of such network dynamics and contextual metrics, we need to cope with the stochastic nature of such environment for allocating uplink and downlink data rates along with gNB association. 
	In the next section, thus, we formalize a decision problem for the virtual space of the XAI twin that can guarantee a zero-touch network and service management and also provide an explained decision for the network service provider.      
	
	\section{Learning Problem Formulation for XAI Twin}
	\label{Learning_Problem_XAI_Twin}
	Given the marginal JPD of network dynamics and contextual metric variables in \eqref{eq:reasoning} and the reasoning of the unobserved variables $\boldsymbol{Z} \in W \setminus \boldsymbol{E}$ in \eqref{eq:most_probable_explanation}, we can determine a \emph{score} for each physical space (i.e., gNB) decisions $(u_{ij}^\textrm{exe}, d_{ij}^\textrm{exe})$ as follows: 
	\begin{equation} \label{eq:Each_gNB_Reward}
		\begin{split}
			\Upsilon(W|a_{ij}^g) = P(\Lambda | (u_{ij}^\textrm{exe}, d_{ij}^\textrm{exe})),
		\end{split}
	\end{equation} 
	where $a_{ij}^g$ represents a candidate gNB association decision in virtual space of XAI twin and $\boldsymbol{a}$ is a vector of all gNB association decisions. 
	We consider a finite time domain $\mathcal{T}$ consisting of $T$ slots with each being indexed by $t$. During $t$, the number of requested IoE service users becomes $|\mathcal{J}_i| |\mathcal{I}|$. Therefore, the expectation of the reasoning score can be determined as follows:   
	\begin{equation} \label{eq:Mean_gNB_Reward}
		\begin{split}
			\mathbb{E}_{P(\Lambda | (u_{ij}^\textrm{exe}, d_{ij}^\textrm{exe}))}[\Gamma(\boldsymbol{a})] =  \sum_{j \in \mathcal{J}_i} \sum_{i \in \mathcal{I}} \Big( \Gamma(a_{ij}^g) + \frac{\Upsilon(W|a_{ij}^g) - \Gamma(a_{ij}^g)}{|\mathcal{J}_i| |\mathcal{I}|}\Big),
		\end{split}
	\end{equation} 	
	where $\Gamma(a_{ij}^g)$ is a current score for the candidate gNB $a_{ij}^g \in \boldsymbol{a}$ while $\Upsilon(W|a_{ij}^g)$ is the reasoning score based on evidence in \eqref{eq:Each_gNB_Reward}. Thus, the updated score of each candidate gNB $a_{ij}^g$ is calculated as follows:
	\begin{equation} \label{eq:Update_Reward_gNB}
		\begin{split}
			\Theta(a_{ij}^g) = \Theta(a_{ij}^g) + \frac{\Upsilon(W|a_{ij}^g) - \Theta(a_{ij}^g)}{(\boldsymbol{\theta}|a_{ij}^g)},
		\end{split}
	\end{equation}
	where $\boldsymbol{\theta}$ represents a knowledge base that is generated from the evidence $\boldsymbol{E}$ in the virtual space of the XAI twin. As a result, we can now devise a learning problem of the XAI twin for reasoning causes, analyzing effects, and executing zero touch network and service management, as follows:
	\begin{subequations}\label{Opt_1_1}
		\begin{align}
			\underset{\boldsymbol{u}, \boldsymbol{d}, \boldsymbol{a}}\min 
			&\;  \frac{1}{T}\sum_{t \in \mathcal{T}}\Big( \mathbb{E}[\Gamma(\boldsymbol{a})] |\mathcal{J}_i| |\mathcal{I}| -  \sum_{j \in \mathcal{J}_i} \sum_{i \in \mathcal{I}} \Theta(a_{ij}^g) \Big) \tag{\ref{Opt_1_1}}, \\
			\text{s.t.} \quad & \label{Opt_1_1:const1} \sum_{\boldsymbol{z} \in \boldsymbol{Z}} P(W_m, \boldsymbol{e}, \boldsymbol{z}) \le \Theta(a_{ij}^g), m \in (1, \dots, M),\\
			&\label{Opt_1_1:const2}  \alpha_{ij}^\textrm{req} \le a_{ij}^g u_{ij}^\textrm{exe} \mathbb{E}[T^\textrm{up}_{ij}], \\
			&\label{Opt_1_1:const3} \beta_{ij}^\textrm{req} \le a_{ij}^g d_{ij}^\textrm{exe} \mathbb{E}[T^\textrm{down}_{ij}], \\
			&\label{Opt_1_1:const4}\chi^\textrm{up} \log_{2}(1+\delta_{ij}^\textrm{phy}) \ge a_{ij}^g u_{ij}^\textrm{exe}, \\
			&\label{Opt_1_1:const5}\chi^\textrm{down} \log_{2}(1+\delta_{ij}^\textrm{phy}) \ge a_{ij}^g d_{ij}^\textrm{exe}, \\
			&\label{Opt_1_1:const6} a_{ij}^g(\upsilon_{1}10 \log_{2} \delta_{ij}^\textrm{phy} + \upsilon_{2}) \ge \eta, \\
			&\label{Opt_1_1:const7} a_{ij}^g\mathbb{E}[\tau_{ij}^\textrm{tot}] \le \tau_{ij}^\textrm{req}, \\
			&\label{Opt_1_1:const8} a_{ij}^g\omega_i \le \omega^\textrm{max}, \\
			& \label{Opt_1_1:const9} a_{ij}^g \in \left\lbrace0,1 \right\rbrace, \forall i \in \mathcal{I}, \forall j \in \mathcal{J}_i,
		\end{align}
	\end{subequations}
	where the learning objective is to minimize the residual between expected explained score \eqref{eq:Mean_gNB_Reward} and current obtained score \eqref{eq:Update_Reward_gNB} on the decision variables of allocated uplink data rate $\forall u_{ij}^\textrm{exe} \in \boldsymbol{u}$, downlink data rate $\forall d_{ij}^\textrm{exe} \in \boldsymbol{d}$, and gNB association $\forall a_{ij}^g \in \boldsymbol{a}$. Constraint \eqref{Opt_1_1:const1} obtains the marginal JPD from the unobserved networks dynamics and contextual features for discretizing probable reasoning on obtained explainable score \eqref{eq:Update_Reward_gNB}. 
	Constraints \eqref{Opt_1_1:const2} and \eqref{Opt_1_1:const3} guarantee that the uplink $u_{ij}^\textrm{exe}$ and downlink $d_{ij}^\textrm{exe}$ data rates will be assigned to IoE user $j \in \mathcal{J}_i$ at gNB $g \in \mathcal{G}$ for a sufficient amount of time such that the amount of uplink $\alpha_{ij}^\textrm{req}$ and downlink $\beta_{ij}^\textrm{req}$ data can be successfully transmitted through the wireless network. 
	In the physical space of XAI twin, for fixed uplink  $\chi^\textrm{up}$ and downlink $\chi^\textrm{down}$ bandwidths, constraints \eqref{Opt_1_1:const4} and \eqref{Opt_1_1:const5} guarantee that the allocated uplink $u_{ij}^\textrm{exe}$ and downlink $u_{ij}^\textrm{exe}$ data rates must be bounded with the systems' capacities. Constraint \eqref{Opt_1_1:const6} ensures a minimum level of CQI $\eta$ for the successful IoE service fulfillment, where $\upsilon_{1} = 0.5223$ and $\upsilon_{2} = 4.6176$ are known coefficients \cite{IEEEhowto:CQI_Li} for CQI measure on SINR $\delta_{ij}^\textrm{phy}$. Constraint \eqref{Opt_1_1:const7} meets the requirement of maximum tolerable delay $\tau_{ij}^\textrm{req}$ for each IoE service user $j \in \mathcal{J}_i$ by characterizing uplink data transfer duration $T^\textrm{up}_{ij}$, waiting time duration $T^\textrm{queue}_{ij}$, execution duration $T^\textrm{exe}_{ij}$, and downlink data transfer duration $T^\textrm{down}_{ij}$ as $\mathbb{E}[\tau_{ij}^\textrm{tot}]$ (in \eqref{eq:j5_tot_exe_delay}) to assign the suitable gNB $a_{ij}^g \in \boldsymbol{a}$. The server utilization rate $\omega_i$ at the assigned gNB $a_{ij}^g \in \boldsymbol{a}$ for IoE service $i \in \mathcal{I}$ must be smaller or equal to its maximum computational capacity $\omega^\textrm{max}$, which is captured by constraint \eqref{Opt_1_1:const8}. Finally, constraint  \eqref{Opt_1_1:const9} guarantees that each IoE service user $j \in \mathcal{J}_i$ is assigned exactly one gNB $a_{ij}^g \in \boldsymbol{a}, g \in \mathcal{G}$ for accomplishing the corresponding service.

	In \eqref{Opt_1_1}, the gNB $a_{ij}^g \in \boldsymbol{a}$ selection depends on the explainable score at the virtual space while the corresponding uplink $u_{ij}^\textrm{exe}$ and downlink $d_{ij}^\textrm{exe}$ data rates allocation completely depend on the network dynamics and contextual metrics of each IoE service request in physical space. 
	Problem \eqref{Opt_1_1} leads to a dynamic programming problem \cite{IEEEhowto:MAB_Book_1} in a stochastic networking environment due to contextual metrics in physical space of each IoE service request is random over time.  
	To solve \eqref{Opt_1_1}, first we estimate uplink $u_{ij}^\textrm{exe}$ and downlink $d_{ij}^\textrm{exe}$ data rates in physical space via a neural-networks-driven multivariate regression model.
	The procedure of finding the uplink $u_{ij}^\textrm{exe}$ and downlink $d_{ij}^\textrm{exe}$ data rates from a pre-determined neural-networks model based on current network dynamics and contextual metrics is called the unconscious decisions into physical space of XAI twin. The unconscious decisions are not aware of the evidence.
	Then, forming a directed acyclic graph-based Bayesian network \cite{IEEEhowto:BN_Base} becomes more appropriate to infer the cause and effects of such decisions through a first-order probabilistic language model \cite{IEEEhowto:Bayesian_Log_Pro, IEEEhowto:FOPL_Base} due to the reasoning score of each gNB $a_{ij}^g \in \boldsymbol{a}$ selection trusted on the marginalized joint probability distribution of all $M$ networks dynamic and contextual metrics.
	Finally, the learning problem \eqref{Opt_1_1} can be reduced to a Bayesian multi-arm bandits \cite{IEEEhowto:MAB_Book_1} as a base problem, where each physical space (i.e., gNB) of the XAI twin can perform the role of a bandit arm. More precisely, we train the virtual space of the XAI twin to reduce the residual between expected explained score and current incurred score to assign a gNB $a_{ij}^g \in \boldsymbol{a}$, given that the unconscious decisions of uplink $u_{ij}^\textrm{exe}$ and downlink $d_{ij}^\textrm{exe}$ data rates for the physical space of XAI twin. A detailed solution procedure for the proposed XAI twin for zero-touch IoE is described in the following section.
	
	\section{Neuro-symbolic XAI Twin Framework Design for Wireless Network}
	\label{NS_XAI_Twin_Framework}
	\begin{figure}[!t]
		\centerline{\includegraphics[scale=.5]{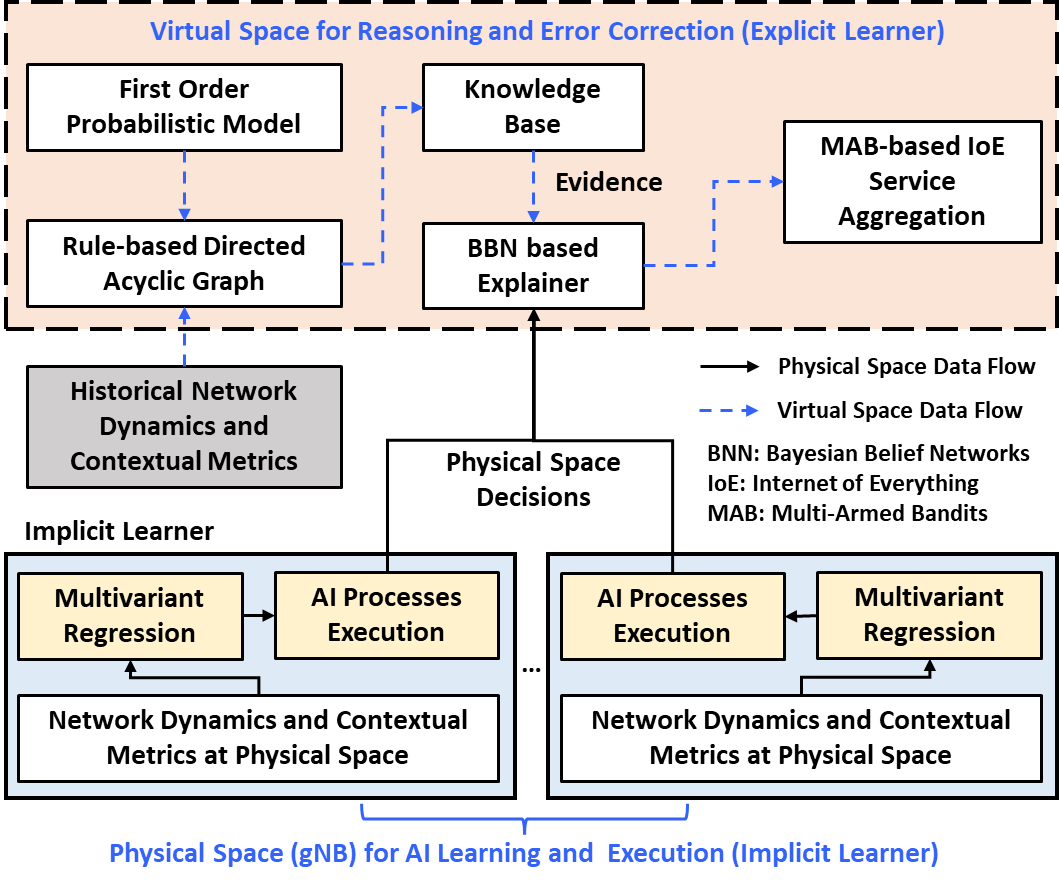}}
		\caption{Holistic view of the proposed neuro-symbolic explainable artificial intelligence twin framework.}
		\label{XAI_Twin_Sol}
	\end{figure}
	The neuro-symbolic XAI twin plays a key role in reasoning the perfect decision of each IoE service execution. In particular, the neuro-symbolic XAI twin has a \emph{neural-networks-driven multivariate regression model for data rate estimation} and the DAG-based Bayesian network with a Bayesian multi-arm bandit \emph{for symbolic reasoning} and IoE service control. 
	On the one hand, the physical space of the XAI-twin at each gNB $g \in \mathcal{G}$ performs the role of an \emph{implicit learner} that can automatically capture the current dynamics of the physical environment and make an unconscious decision. On the other hand, the virtual space at the $5$G core executes exploitation by symbolic reasoning of the physical space decisions and prior knowledge for error correction by an \emph{explicit leaner}. A holistic view of the proposed neuro-symbolic explainable artificial intelligence twin framework is shown in Figure \ref{XAI_Twin_Sol}.
	
	\subsection{Implicit Learner of XAI Twin}
	\label{ImplicitLearningofXAITwin}
	A tuple $X \coloneqq (s_{ij}^\textrm{phy}, p_{ij}^\textrm{phy}, q_{ij}^\textrm{phy}, \delta_{ij}^\textrm{phy}, c_{ij}^\textrm{phy})$ of contextual metrics at a physical space (i.e., gNB) will play the key role for allocating uplink $u_{ij}^\textrm{exe}$ and downlink $d_{ij}^\textrm{exe}$ data transfer rates to each IoE user. Here, it is necessary to meet the IoE service requirements such as amount of upload $\alpha_{ij}^\textrm{req}$ (in \eqref{Opt_1_1:const2} and \eqref{Opt_1_1:const4}), download $\beta_{ij}^\textrm{req}$ data (in \eqref{Opt_1_1:const3} and \eqref{Opt_1_1:const5}), maintaining CQI (in \eqref{Opt_1_1:const6}), end to end communication and computational delay  $\tau_{ij}^\textrm{req}$ (in \eqref{Opt_1_1:const7}), and utilization of computational capacity $\omega_i$ (in \eqref{Opt_1_1:const8}) become crucial on contextual metrics $X$ at the physical space. 
	On top of the proposed implicit learner, we can flexibly deploy numerous regression algorithms \cite{IEEEhowto:LSTM_Base, IEEEhowto:sklearn_ensemble} such as long short-term memory (LSTM), deep neural network-base regression, extra Trees, random forest, adaboost, and logistic regression based on the service provider requirements. However, the challenge is to determine a maximum likelihood estimation of $\boldsymbol{\xi}$ in \eqref{eq:Mul_Var_Reg}. Thus, we design a least square estimator $\hat{\boldsymbol{\xi}}$ \cite{IEEEhowto:Multivariate_Regression_1, IEEEhowto:Multivariate_Regression_2, IEEEhowto:Multivariate_Regression_3} of \eqref{eq:Mul_Var_Reg} as follows:
	\begin{equation} \label{eq:Phy_least_square_estimation}
		\begin{split}
			\hat{\boldsymbol{\xi}} =(\boldsymbol{X}^{\textrm{T}}\boldsymbol{X})^{-1} \boldsymbol{X}^{\textrm{T}} \boldsymbol{Y},
		\end{split}
	\end{equation}
	where $u_{ij}^\textrm{exe}, d_{ij}^\textrm{exe} \in \boldsymbol{Y}$ and $X \in \boldsymbol{X}$. Therefore, $(\boldsymbol{Y} - \boldsymbol{X}\hat{\boldsymbol{\xi}})^{\textrm{T}} (\boldsymbol{Y} - \boldsymbol{X}\hat{\boldsymbol{\xi}})$ is a learning objective of \eqref{eq:Phy_least_square_estimation}.
	The \eqref{eq:Phy_least_square_estimation} becomes a least square minimization problem for an estimator $\hat{\boldsymbol{\xi}}$. However, the dynamics of each IoE user $j \in \mathcal{J}_i$ is heterogeneous in nature based on observed contextual metrics $X$. As a result, the estimator $\hat{\boldsymbol{\xi}}$ belongs to an biased decision. Thus, to eliminate the bias of the estimator $\hat{\boldsymbol{\xi}}$, utilization of a covariance matrix $\boldsymbol{\Omega}$ of $\boldsymbol{\epsilon}$ (in \eqref{eq:Mul_Var_Reg}) is more suitable. Therefore, an unbiased estimator $\hat{\boldsymbol{\Omega}}$ can be defined as follows:
	\begin{equation} \label{eq:Phy_unbiased_estimation}
		\begin{split}
			\hat{\boldsymbol{\Omega}} = \boldsymbol{\Omega}(\boldsymbol{Y} - \boldsymbol{X}\hat{\boldsymbol{\xi}})^{\textrm{T}} (\boldsymbol{Y} - \boldsymbol{X}\hat{\boldsymbol{\xi}}),
		\end{split}
	\end{equation}
	where $\boldsymbol{\Omega} = \frac{1}{(J-N-1)}$ represents a covariance matrix with a zero mean. Then the implicit learner of XAI twin fits the for the unconscious decision as follows: 
	\begin{equation} \label{eq:Phy_fit_model}
		\begin{split}
			\hat{\boldsymbol{Y}} = \boldsymbol{X} \hat{\boldsymbol{\xi}},
		\end{split}
	\end{equation}
	where estimation error become $\hat{\boldsymbol{\epsilon}} = \boldsymbol{Y} - \hat{\boldsymbol{Y}}$. We illustrate the working procedure of the proposed implicit learner in Algorithm \ref{alg:implicit_learner}. 
	
	\begin{algorithm}[t!]
		\caption{Implicit Learner at Each gNB $g \in \mathcal{G}$}
		\label{alg:implicit_learner}
		\begin{algorithmic}[1]
			\renewcommand{\algorithmicrequire}{\textbf{Input:}}
			\renewcommand{\algorithmicensure}{\textbf{Output:}}
			\REQUIRE  $ \forall X \coloneqq (s_{ij}^\textrm{phy}, p_{ij}^\textrm{phy}, q_{ij}^\textrm{phy}, \delta_{ij}^\textrm{phy}, c_{ij}^\textrm{phy}) \in \boldsymbol{X}$, $\forall (u_{ij}^\textrm{req}, d_{ij}^\textrm{req}) \in \boldsymbol{Y}$, $\forall ($ $\alpha_{ij}^\textrm{req}$, $\beta_{ij}^\textrm{req}$, $\tau_{ij}^\textrm{req}$$) \in  \mathcal{J}_i, \mathcal{I}$
			\ENSURE  $\forall (u_{ij}^\textrm{exe}, d_{ij}^\textrm{exe}) \in \hat{\boldsymbol{Y}}, \in \mathcal{J}_i, \mathcal{I}$, $W$ 
			\\ \textit{Initialization:} $W$, $\omega^\textrm{max}$, $\eta$, $\chi^\textrm{down}$, $\chi^\textrm{up}$, $\boldsymbol{\Omega}$, $\lambda_i$, $\nu$, \textit{regression parameters based on specific regression algorithms}
			\WHILE{($\forall X \in \boldsymbol{X}$)}
			\STATE \textbf{Estimate} $\hat{\boldsymbol{\xi}}$ \textbf{:} using unbiased least square minimization objective $\boldsymbol{\Omega}(\boldsymbol{Y} - \boldsymbol{X}\hat{\boldsymbol{\xi}})^{\textrm{T}} (\boldsymbol{Y} - \boldsymbol{X}\hat{\boldsymbol{\xi}})$ in \eqref{eq:Phy_unbiased_estimation}
			\STATE \textbf{Model fit} $\hat{\boldsymbol{Y}}$\textbf{:} $\boldsymbol{X} \hat{\boldsymbol{\xi}}$ using \eqref{eq:Phy_fit_model}, 
			$\forall (u_{ij}^\textrm{exe}, d_{ij}^\textrm{exe}) \in \hat{\boldsymbol{Y}}$
			\STATE \textbf{Evaluate:} $\hat{\boldsymbol{\epsilon}} = \boldsymbol{Y} - \hat{\boldsymbol{Y}}$
			\IF {($\hat{\boldsymbol{\epsilon}} \le \nu$)}
			\STATE\textbf{Break}
			\ENDIF
			\ENDWHILE
			\WHILE{$\forall i \in \mathcal{I}$} 
			\FOR {$\forall j \in \mathcal{J}_i$ }
			\STATE \textbf{Mean uplink duration} $\mathbb{E}[T^\textrm{up}_{ij}]$\textbf{:} $\sum_{\forall j \in \mathcal{J}_i} \tau_{ij}^{\textrm{up}} P(\tau_{ij}^{\textrm{up}})$
			\STATE \textbf{Service computational rate} $\mu_i$\textbf{:} $\frac{\lambda_i \alpha_{ij}^\textrm{req}}{\gamma_i}$
			\STATE \textbf{Estimate server utilization rate} $\omega_i$\textbf{:} $\sum_{\forall j \in \mathcal{J}_i} \frac{\lambda_i}{\mu_i}$
			\STATE \textbf{Mean waiting duration} $\mathbb{E}[T^\textrm{queue}_{ij}]$\textbf{:} $\frac{\lambda_i \mathbb{E}[(T_{ij}^\textrm{service})^2]}{2(1-\omega_i)}$
			\STATE \textbf{Mean computational and waiting duration:} $\mathbb{E}[T^\textrm{exe}_{ij}] = (1+ \frac{1+C^2}{2} \frac{\omega_i}{1-\omega_i})\mathbb{E}[T_{ij}^\textrm{service}]$ 
			\STATE \textbf{Mean downlink duration} $\mathbb{E}[T^\textrm{down}_{ij}]$\textbf{:} $\sum_{\forall j \in \mathcal{J}_i} \tau_{ij}^{\textrm{down}} P(\tau_{ij}^{\textrm{down}})$
			\STATE \textbf{Calculate mean of total delay} $\mathbb{E}[\tau_{ij}^\textrm{tot}]$\textbf{:} $\mathbb{E}[T^\textrm{up}_{ij}] + \mathbb{E}[T^\textrm{queue}_{ij}] + \mathbb{E}[T^\textrm{exe}_{ij}] + \mathbb{E}[T^\textrm{down}_{ij}],$
			\IF {(Satisfy constraints \eqref{Opt_1_1:const2}, \eqref{Opt_1_1:const3}, \eqref{Opt_1_1:const4}, \eqref{Opt_1_1:const5}, \eqref{Opt_1_1:const6}, \eqref{Opt_1_1:const7}, and \eqref{Opt_1_1:const8})}
			\STATE\textbf{Assign:} $a_{ij}^g =1$ 
			\ELSE
			\STATE\textbf{Assign:} $a_{ij}^g =0$ 
			\ENDIF
			\STATE\textbf{Concatenation:} $W= X + u_{ij}^\textrm{exe}, d_{ij}^\textrm{exe}, a_{ij}^g$
			\ENDFOR
			\ENDWHILE
			\RETURN $ (s_{ij}^\textrm{phy}, p_{ij}^\textrm{phy}, q_{ij}^\textrm{phy}, \delta_{ij}^\textrm{phy}, c_{ij}^\textrm{phy}, u_{ij}^\textrm{exe}, d_{ij}^\textrm{exe}, a_{ij}^g) \in W, \forall \mathcal{J}_i, \mathcal{I}$  
		\end{algorithmic} 
	\end{algorithm} 
	
	Each gNB $g \in \mathcal{G}$ is responsible to execute the proposed \emph{implicit learner} Algorithm \ref{alg:implicit_learner}. The role of the Algorithm \ref{alg:implicit_learner} is to make an unconscious decisions $u_{ij}^\textrm{exe}, d_{ij}^\textrm{exe}, a_{ij}^g$ based on current contextual metrics $ X \coloneqq (s_{ij}^\textrm{phy}, p_{ij}^\textrm{phy}, q_{ij}^\textrm{phy}, \delta_{ij}^\textrm{phy}, c_{ij}^\textrm{phy})$ and each IoE service requirements $u_{ij}^\textrm{req}$, $d_{ij}^\textrm{req}$, $\alpha_{ij}^\textrm{req}$, $\beta_{ij}^\textrm{req}$, $\tau_{ij}^\textrm{req}$ at each gNB $g \in \mathcal{G}$. Lines from $1$ to $8$ in Algorithm \ref{alg:implicit_learner} are responsible to execute a multivariate regression \cite{IEEEhowto:Multivariate_Regression_1, IEEEhowto:Multivariate_Regression_2, IEEEhowto:Multivariate_Regression_3} scheme, where numerous regression process \cite{IEEEhowto:LSTM_Base, IEEEhowto:sklearn_ensemble} can be flexibly deployed based on the service provider demands. 
	Thus, we estimate $\hat{\boldsymbol{\xi}}$ using the proposed unbiased least square minimization objective \eqref{eq:Phy_fit_model} at line $2$ in the Algorithm \ref{alg:implicit_learner}. Lines $3$ and $4$ fit the estimator $\hat{\boldsymbol{\xi}}$ to predict  $(u_{ij}^\textrm{exe}, d_{ij}^\textrm{exe}) \in  \hat{\boldsymbol{Y}}$ and evaluate the prediction error $\hat{\boldsymbol{\epsilon}}$, respectively. A termination decision of the regression process is made in line $5$ of Algorithm \ref{alg:implicit_learner} based on the required tolerable prediction error $\nu$. 
	The convergence of regression process depends on the value of parameter $\hat{\boldsymbol{\epsilon}} \le \nu$ \cite{IEEEhowto:Multivariate_Regression_1, IEEEhowto:Multivariate_Regression_2, IEEEhowto:Multivariate_Regression_3} and number of iterations relies on parameter settings of specific regression algorithms \cite{IEEEhowto:LSTM_Base, IEEEhowto:sklearn_ensemble} such as long short-term memory (LSTM), deep neural network-base regression, extra Trees, random forest, adaboost, and logistic regression \cite{IEEEhowto:LSTM_Base, IEEEhowto:sklearn_ensemble}. 
	
	An expectation of total IoE service fulfillment delay $\mathbb{E}[\tau_{ij}^\textrm{tot}]$ is estimated in Algorithm \ref{alg:implicit_learner} at line $16$. This delay is then combined with uplink duration $\mathbb{E}[T^\textrm{up}_{ij}]$, waiting time duration $\mathbb{E}[T^\textrm{queue}_{ij}]$, computational duration $\mathbb{E}[T^\textrm{exe}_{ij}]$, and downlink duration $\mathbb{E}[T^\textrm{down}_{ij}]$ that are calculated in lines $11$, $14$, $15$, and $16$ of Algorithm \ref{alg:implicit_learner}, respectively. Line $18$ in Algorithm \ref{alg:implicit_learner} is responsible to evaluate constraints \eqref{Opt_1_1:const2}, \eqref{Opt_1_1:const3}, \eqref{Opt_1_1:const4}, \eqref{Opt_1_1:const5}, \eqref{Opt_1_1:const6}, \eqref{Opt_1_1:const7}, and \eqref{Opt_1_1:const8} based on the implicit decisions of $u_{ij}^\textrm{exe}$, and $d_{ij}^\textrm{exe}$ at gNB $g \in \mathcal{G}$. An decision indicator $a_{ij}^g$ is marked in lines $19$ or $21$ based on the constraints satisfaction of line $18$ in Algorithm \ref{alg:implicit_learner}. Finally, network dynamics and contextual metrics are preserved in line $23$ (in Algorithm \ref{alg:implicit_learner}) for reasoning and error correction by utilizing these findings in the virtual space of the XAI twin. 
	
	The output of Algorithm \ref{alg:implicit_learner} includes unconscious decisions $u_{ij}^\textrm{exe}, d_{ij}^\textrm{exe}, a_{ij}^g$ based on current contextual metrics $ X \coloneqq (s_{ij}^\textrm{phy}, p_{ij}^\textrm{phy}, q_{ij}^\textrm{phy}, \delta_{ij}^\textrm{phy}, c_{ij}^\textrm{phy})$ for all IoE service $\forall \mathcal{J}_i, \mathcal{I}$ at gNB $g \in \mathcal{G}$. Thus, Algorithm \ref{alg:implicit_learner} at physical space determines $W= X + u_{ij}^\textrm{exe}, d_{ij}^\textrm{exe}, a_{ij}^g$ that will use for symbolic reasoning by a building Bayesian DAG. The complexity of the proposed \emph{implicit learner} Algorithm \ref{alg:implicit_learner} belongs to $\mathcal{O}(|\boldsymbol{X}|+ |\mathcal{I}||\mathcal{J}_i|)$, where $|\boldsymbol{X}|$, $|\mathcal{I}|$, and $|\mathcal{J}_i|$ are size of contextual metrics, number of IoE service by the service provider, and number of user for each IoE service $j \in \mathcal{J}_i$ respectively. However, the complexity of the multivariate regression $\mathcal{O}(|\boldsymbol{X}|)$ \cite{IEEEhowto:Multivariate_Regression_1, IEEEhowto:Multivariate_Regression_2, IEEEhowto:Multivariate_Regression_3} can vary depending on the algorithmic procedures of several schemes and neural network configurations \cite{IEEEhowto:LSTM_Base, IEEEhowto:sklearn_ensemble}. In the following section, we present explicit learner of XAI twin for reasoning and error correction on implicit learners' decisions for assuring ZSM.
	
	\subsection{Explicit Learner of XAI Twin}
	\label{ExplicitLearningofXAITwin}
	The role of an \emph{explicit learner} in our XAI twin is to find the root causes behind the physical space decisions for each gNB $g \in \mathcal{G}$ and perform autonomous error correction based on the reasoning.
	However, the interpretation of reasoning and error correction become challenging since the network dynamics and contextual metrics are nondeterministic for all physical space $\forall g \in \mathcal{G}$ due to each IoE service includes indispensable characteristics such as user, data, processes, and things \cite{IEEEhowto:Manogaran_IoE_2, IEEEhowto:Adhikari_IoE_4, IEEEhowto:Pant_IoE_3}. Thus, to design an \emph{explicit learner}, a symbolic reasoning approach \cite{IEEEhowto:NS_coherence_2System, IEEEhowto:Guo_XAI_1} can be more appropriate for achieving higher accuracy, data efficiency, transparent, and trustworthy decisions towards the ZSM \cite{IEEEhowto:Chergui_AI_Zero, IEEEhowto:ETSI_Zero_Touch, IEEEhowto:5G_Architecture_3PPP}. We design the explicit learner of the XAI twin by incorporating a first-order probabilistic language model \cite{IEEEhowto:Bayesian_Log_Pro, IEEEhowto:FOPL_Base} for reasoning the network dynamics and contextual metrics, and we devise a Bayesian multi-arm bandits system \cite{IEEEhowto:MAB_Book_1} for making autonomous corrected decisions based on the reasoning.     
	
	\begin{figure}[!t]
		\centerline{\includegraphics[scale=.63]{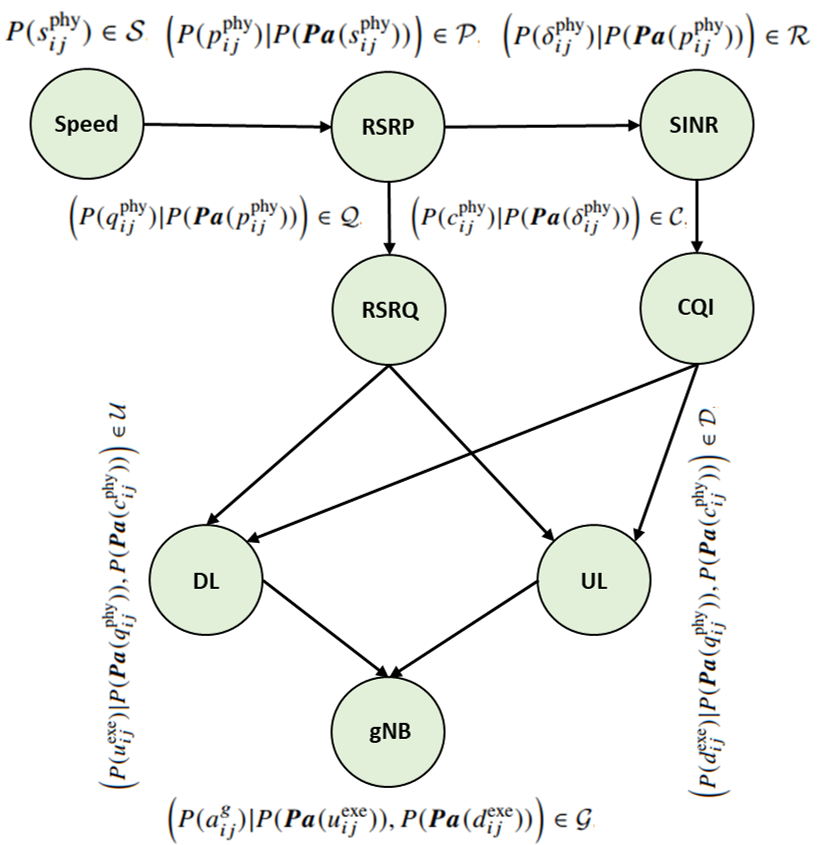}}
		\caption{Logical reasoning by encoding influences among the network dynamics and contextual metrics using a directed acyclic graph for the explicit learner.}
		\label{DAG_Fig}
	\end{figure}
	
	We consider a DAG \cite{IEEEhowto:Bayesian_Log_Pro} as seen in Figure \ref{DAG_Fig}, where a set of rules is defined by the service provider for each network dynamic and contextual metric $W_m$ (i.e., node in DAG) to meet the IoE service requirements. For instance, we consider $\mathcal{S}$, $\mathcal{P}$, $\mathcal{Q}$, $\mathcal{R}$, $\mathcal{C}$, $\mathcal{U}$, $\mathcal{D}$, and $\mathcal{G}$ as the sets of rules for observing user speed $s_{ij}^\textrm{phy}$, RSRP $p_{ij}^\textrm{phy}$, RSRQ $q_{ij}^\textrm{phy}$, SINR $\delta_{ij}^\textrm{phy}$, CQI $c_{ij}^\textrm{phy}$, uplink data rate $u_{ij}^\textrm{exe}$, downlink data rate $d_{ij}^\textrm{exe}$, and user association of gNB $a_{ij}^g$, respectively. Thus, we can characterize unobserved network dynamics and contextual metric variables $\forall \boldsymbol{z} \in \boldsymbol{Z}$ for \eqref{eq:reasoning}, where $\boldsymbol{Z} \in W \setminus \boldsymbol{E}$. A Bayesian-based \cite{IEEEhowto:BN_Base} first-order probabilistic language model \cite{IEEEhowto:Bayesian_Log_Pro, IEEEhowto:FOPL_Base} allows us to estimate a tuple of random variables $W = (W_1, W_2, \dots, W_M)$ for reasoning network dynamics and contextual parameters as follows:     
	\begin{subequations}\label{KB_BBN_DAG}
		\begin{align}
			&\; \sum_{\boldsymbol{z} \in \boldsymbol{Z}} P(W_m, \boldsymbol{e}, \boldsymbol{z}), \forall m \in M \tag{\ref{KB_BBN_DAG}}, \\
			\text{nodes:} \quad & \label{KB_BBN_DAG:const1} P(s_{ij}^\textrm{phy}) \in \mathcal{S},\\
			&\label{KB_BBN_DAG:const2}  \Big(P(p_{ij}^\textrm{phy}) | P(\boldsymbol{Pa}(s_{ij}^\textrm{phy}))\Big) \in \mathcal{P}, \\
			&\label{KB_BBN_DAG:const3} \Big(P(\delta_{ij}^\textrm{phy}) | P(\boldsymbol{Pa}(p_{ij}^\textrm{phy}))\Big) \in \mathcal{R}, \\
			&\label{KB_BBN_DAG:const4}	\Big(P(q_{ij}^\textrm{phy}) | P(\boldsymbol{Pa}(p_{ij}^\textrm{phy}))\Big) \in \mathcal{Q}, \\
			&\label{KB_BBN_DAG:const5}	\Big(P(c_{ij}^\textrm{phy}) | P(\boldsymbol{Pa}(\delta_{ij}^\textrm{phy}))\Big) \in \mathcal{C}, \\
			&\label{KB_BBN_DAG:const6}	\Big(P(d_{ij}^\textrm{exe}) | P(\boldsymbol{Pa}(q_{ij}^\textrm{phy})), P(\boldsymbol{Pa}(c_{ij}^\textrm{phy}))\Big) \in \mathcal{D}, \\
			&\label{KB_BBN_DAG:const7} \Big(P(u_{ij}^\textrm{exe}) | P(\boldsymbol{Pa}(q_{ij}^\textrm{phy})), P(\boldsymbol{Pa}(c_{ij}^\textrm{phy}))\Big) \in \mathcal{U}, \\
			&\label{KB_BBN_DAG:const8} \Big(P(a_{ij}^g) | P(\boldsymbol{Pa}(u_{ij}^\textrm{exe})), P(\boldsymbol{Pa}(d_{ij}^\textrm{exe}))\Big) \in \mathcal{G}, 
		\end{align}
	\end{subequations}
	where \eqref{KB_BBN_DAG} determines a marginalized JPD over the unobserved network dynamics and contextual metrics using \eqref{eq:most_probable_explanation} while from \eqref{KB_BBN_DAG:const1} to \eqref{KB_BBN_DAG:const8} represent the nodes of the considered DAG. 
	\eqref{KB_BBN_DAG:const1}-\eqref{KB_BBN_DAG:const8} can determine the marginal JPD of $M$ network dynamics and contextual parameters $W = (W_1, W_2, \dots, W_M)$ for reasoning. The left side of the constraint \eqref{Opt_1_1:const1} can be captured by \eqref{KB_BBN_DAG:const1}-\eqref{KB_BBN_DAG:const8} while \eqref{eq:Update_Reward_gNB} can determine the right side of \eqref{Opt_1_1:const1}. 
	To learn the objective of \eqref{Opt_1_1}, we can redefine it as a function of expected score \eqref{eq:Mean_gNB_Reward} given that the observed evidence of the physical space decisions on uplink $u_{ij}^\textrm{exe}$ and downlink $d_{ij}^\textrm{exe}$ data rates for each IoE service user $j \in \mathcal{J}_i$. Therefore, for a gNB selection decision variable $a_{ij}^g \in \boldsymbol{a}$, the residual of reasoning for the explicit learner is defined as follows:
	\begin{equation} \label{eq:Bay_fn_regret}
		\begin{split}
			\Phi(\mathbb{E}[\Gamma(\boldsymbol{a})|\boldsymbol{Z}]) = \;\;\;\;\;\;\;\;\;\;\;\;\;\;\;\;\;\;\;\;\; \\ \underset{\forall a_{ij}^g \in \boldsymbol{a}} \min \; \mathbb{E}_{\boldsymbol{Z} \sim P(\Lambda)} \Big[ \mathbb{E}[\Gamma(\boldsymbol{a})] |\mathcal{J}_i| |\mathcal{I}| -  \sum_{j \in \mathcal{J}_i} \sum_{i \in \mathcal{I}} \Theta(a_{ij}^g) \Big].
		\end{split}
	\end{equation}
	Each physical space of the XAI twin is considered as a bandit and we can select a gNB $a_{ij}^g \in \boldsymbol{a}$ for fulfilling each IoE service. Thus, the exploration can be determined by the well-known upper confidence bound (UCB) bandit scheme since it can handle the fact of uncertainty for the reasoning \cite{IEEEhowto:MAB_Book_1} and the exploration mechanism for explicit learner is given as follows:
	\begin{equation} \label{eq:gNB_selection}
		\begin{split}
			a_{ij}^g = \underset{a_{ij}^g \in \boldsymbol{a}} \argmax \; \Big(\Theta(a_{ij}^g) + \phi \sqrt{\frac{\log(j)}{\boldsymbol{\theta}}} \Big),
		\end{split}
	\end{equation} 
	where $\phi$ exploration coefficient, $j$ is the index of current IoE user, and $\boldsymbol{\theta}$ represents the knowledge base of all user $\forall \theta_{i,j}^{g} \in \boldsymbol{\theta}$. We summarize the working procedure of the proposed explicit learner in Algorithm \ref{alg:explicit_learner}. 
	\begin{algorithm}[t!]
		\caption{Explicit Learner at Virtual Space of XAI Twin}
		\label{alg:explicit_learner}
		\begin{algorithmic}[1]
			\renewcommand{\algorithmicrequire}{\textbf{Input:}}
			\renewcommand{\algorithmicensure}{\textbf{Output:}}
			\REQUIRE  $ (s_{ij}^\textrm{phy}, p_{ij}^\textrm{phy}, q_{ij}^\textrm{phy}, \delta_{ij}^\textrm{phy}, c_{ij}^\textrm{phy}, u_{ij}^\textrm{exe}, d_{ij}^\textrm{exe}, a_{ij}^g) \in W, \forall \mathcal{J}_i, \mathcal{I}$
			\ENSURE  $\boldsymbol{z} \in \boldsymbol{Z}$, $\forall u_{ij}^\textrm{exe} \in \boldsymbol{u}$, $\forall d_{ij}^\textrm{exe} \in \boldsymbol{d}$, $\forall a_{ij}^g \in \boldsymbol{a}$.
			\\ \textit{Initialization:}  $\mathcal{S}$, $\mathcal{P}$, $\mathcal{Q}$, $\mathcal{R}$, $\mathcal{C}$, $\mathcal{U}$, $\mathcal{D}$, $\mathcal{G}$, $\phi$, $\forall W \in \mathcal{W}$ 
			\FOR {$\forall g \in \mathcal{G}$ }
			\STATE \textbf{Collect all:} $\forall W \in \mathcal{W}$,  $t \in \mathcal{T}$  
			\FOR {$\forall m \in W \in \mathcal{W} $ }
			\STATE \textbf{Estimate:} $P(W_m|\boldsymbol{e})$ using \eqref{eq:reasoning}
			\STATE \textbf{Probable explanation:} $P(\Lambda)$ using \eqref{eq:most_probable_explanation}
			\IF {(\eqref{KB_BBN_DAG:const1}-\eqref{KB_BBN_DAG:const8})}
			\STATE \textbf{Calculate:} $\sum_{\boldsymbol{z} \in \boldsymbol{Z}} P(W_m, \boldsymbol{e}, \boldsymbol{z})$
			\ENDIF
			\ENDFOR		
			\STATE \textbf{Append:} $\forall \theta_{i,j}^{g} \in \boldsymbol{\theta}$
			\ENDFOR
			\FOR {$\forall t \in \mathcal{T}$ }
			\FOR {Until $|\mathcal{G}||\mathcal{I}||\mathcal{J}_i|$ }
			\STATE \textbf{Estimate each score:} $P(\Lambda | (u_{ij}^\textrm{exe}, d_{ij}^\textrm{exe}))$ using \eqref{eq:Each_gNB_Reward}
			\STATE \textbf{Estimate mean score} $\mathbb{E}[\Gamma(\boldsymbol{a})]$ using \eqref{eq:Mean_gNB_Reward}, 
			\STATE \textbf{Find gNB:} $a_{ij}^g = \underset{a_{ij}^g \in \boldsymbol{a}} \argmax \; \Big(\Theta(a_{ij}^g) + \phi \sqrt{\frac{\log(j)}{\boldsymbol{\theta}}} \Big)$
			\STATE \textbf{Update current score:} $\Theta(a_{ij}^g)$ using \eqref{eq:Update_Reward_gNB}
			\ENDFOR
			\STATE \textbf{Evaluate residual score:} $\Phi(\mathbb{E}[\Gamma(\boldsymbol{a})|\boldsymbol{Z}])$ using \eqref{eq:Bay_fn_regret}
			\ENDFOR
			\RETURN $\boldsymbol{z} \in \boldsymbol{Z}$, $\forall u_{ij}^\textrm{exe} \in \boldsymbol{u}$, $\forall d_{ij}^\textrm{exe} \in \boldsymbol{d}$, $\forall a_{ij}^g \in \boldsymbol{a}$ 
		\end{algorithmic} 
	\end{algorithm}
	
	The role of Algorithm \ref{alg:explicit_learner} is to determine a conscious decision for assuring ZSM with reliability. The input of the Algorithm \ref{alg:explicit_learner} is an output of Algorithm \ref{alg:implicit_learner} which includes the network dynamics and contextual metrics $W$ based on implicit decision by Algorithm \ref{alg:implicit_learner}. In Algorithm \ref{alg:explicit_learner}, lines form $1$ to $11$, we determine the probable explanation score based on evidence and symbolic reasoning. In particular, line $5$ determines the probable explanation while line $6$ validates the symbolic reasoning using \eqref{KB_BBN_DAG:const1}-\eqref{KB_BBN_DAG:const8} in Algorithm \ref{alg:explicit_learner}. We learn the system from lines $12$ to $19$ in Algorithm \ref{alg:explicit_learner}. In Algorithm \ref{alg:explicit_learner}, lines $14$, $15$, $16$, and $17$ are responsible for devising each explainable score, mean score, gNB selection, and updated  explainable score, respectively, for each IoE service fulfillment decision. Finally, residual score is evaluated in line $19$ of Algorithm \ref{alg:explicit_learner} for determining  $\boldsymbol{z} \in \boldsymbol{Z}$, $\forall u_{ij}^\textrm{exe} \in \boldsymbol{u}$, $\forall d_{ij}^\textrm{exe} \in \boldsymbol{d}$, $\forall a_{ij}^g \in \boldsymbol{a}$ to assure a ZSM in wireless network. The worst-case execution complexity of Algorithm \ref{alg:explicit_learner} leads to $\mathcal{O}((|\mathcal{I}||\mathcal{J}_i| \log |\mathcal{I}||\mathcal{J}_i|) + M^2 )$, where $|\mathcal{I}|$, $|\mathcal{J}_i|$, and $M$ are the number of IoE services, number of IoE service sessions, and number of DAG nodes.    
	
	\section{Experimental Results and Discussion}
	\label{Experimental_Result}
	\begin{table}[t!]
		\caption{Summary of Experiment Setup }
		\begin{center}
			\begin{tabular}{|p{2.5cm}|p{5.5cm}|}
				\hline
				\textbf{Description}&{\textbf{Values}} \\
				\hline
				Set of gNB $\mathcal{S}$ & $\left\{{1,\cdots, 5}\right\}$ \cite{IEEEhowto:Data_set_5g}\\
				\hline
				No. of IoE sessions $|\mathcal{I}| |\mathcal{J}_i|$  & $2206$ \cite{IEEEhowto:Data_set_5g}\\
				\hline
				Set of speed range $\mathcal{S}$ & $\left\{>=80, 60-80, 30-60, <=30\right\}$ [km/h] \cite{IEEEhowto:Data_set_5g}\\
				\hline
				Set of RSRP range $\mathcal{P}$ & $\left\{>=-80, -80-(-90), -90-(-100), <=-100\right\}$ [dBm] \cite{IEEEhowto:Data_set_5g}\\
				\hline
				Set of RSRQ range $\mathcal{Q}$ & $\left\{>=-10, (-10)-(-15), <=-15\right\}$ [dB] \cite{IEEEhowto:Data_set_5g}\\
				\hline
				Set of SINR range $\mathcal{R}$ & $\left\{>=20, 13-20, 0-13, <=0\right\}$ [dB] \cite{IEEEhowto:Data_set_5g}\\
				\hline
				Set of CQI range $\mathcal{C}$ & $\left\{>=10, 7-10, <=7\right\}$  \cite{IEEEhowto:Data_set_5g}\\
				\hline
				Set of uplink data rate range $\mathcal{U}$ & $\left\{>=100000, 50000-100000, <=50000\right\}$ [Kbps] \cite{IEEEhowto:Data_set_5g}\\
				\hline
				Set of downlink data rate range $\mathcal{D}$ & $\left\{>=800, 100-800, <=100\right\}$ [Kbps] \cite{IEEEhowto:Data_set_5g}\\
				\hline
				Downloaded data size& $>200$ Mb \cite{IEEEhowto:Data_set_5g} \\
				\hline
				Wider range bandwidth & $20$ MHz \cite{IEEEhowto:3GPP_1}\\
				\hline
				No. of sub-carriers  & $12$ \cite{IEEEhowto:3GPP_1}\\
				\hline
				Carrier frequency of each resource block & $180$ KHz \cite{IEEEhowto:3GPP_1}\\
				\hline
				RBs at per channel bandwidth  & $100$ \cite{IEEEhowto:3GPP_1}\\
				\hline
				Phy. space pertraining epochs & $5000$ \\
				\hline
				Phy. space optimizer & ADAM \\
				\hline
				Phy. space LSTM units  & $128$ \\
				\hline
				Phy. space activation  & ReLU \\
				\hline
				No. of random variables in virtual space $|M|$& $7$ \\
				\hline
				No. of iteration in virtual space & $200$ \\
				\hline
				Exploration coefficient $\phi$  & $1$\\
				\hline
				Virtual space KB sessions & $662$\\
				\hline
				XAI twin testing sessions & $100$, $205$\\
				\hline
			\end{tabular}
			\label{tab2_sim_param}
		\end{center}
	\end{table}
	For our experiments, we consider a state-of-the-art $5$G dataset (B$\_2020.02.13\_13.03.24$) \cite{IEEEhowto:Data_set_5g} and important parameters are shown is Table \ref{tab2_sim_param}. We have implemented both Algorithms \ref{alg:implicit_learner} and \ref{alg:explicit_learner} in Python \cite{IEEEhowto:keras, IEEEhowto:sklearn_ensemble, IEEEhowto:PyBBN} over a desktop environment with a core i$9$ processor and $64$ GB random access memory. In fact, in case of real deployment, the implicit learner \ref{alg:implicit_learner} and explicit learner \ref{alg:explicit_learner} will be deployed in each gNB (i.e., physical space) and EPC core (i.e., virtual space), respectively. We benchmark the proposed neuro-symbolic XAI twin by comparing it with a theoretical analogy similar to the considered dataset's (B$\_2020.02.13\_13.03.24$) \cite{IEEEhowto:Data_set_5g} ground truth and several XAI-supported ensemble-based regression schemes  \cite{IEEEhowto:sklearn_ensemble} such as Random Forest, Extra Trees, AdaBoost, and Linear Regression. Further, we have compared the proposed neuro-symbolic XAI twin with the LSTM and deep neural networks-based regression model in terms of accuracy for IoE service execution. We validate the trustworthiness score of the proposed neuro-symbolic XAI twin by comparing it with Gradient-based bandits and Epsilon-greedy \cite{IEEEhowto:MAB_Book_1} mechanisms.

	\subsection{Performance and Reliability of the Proposed Neuro-symbolic XAI Twin}
	\begin{figure}[!t]
		\centerline{\includegraphics[width=8.3cm]{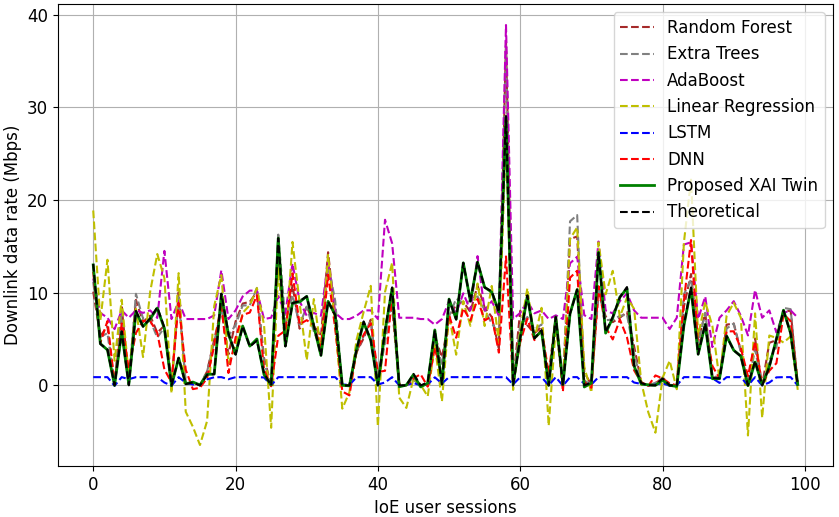}}
		\caption{Comparison of the allocated downlink data rate of the proposed neuro-symbolic XAI twin with other baselines.}
		\label{Downlink_data_rate}
	\end{figure}
	\begin{figure}[!t]
		\centerline{\includegraphics[width=8.3cm]{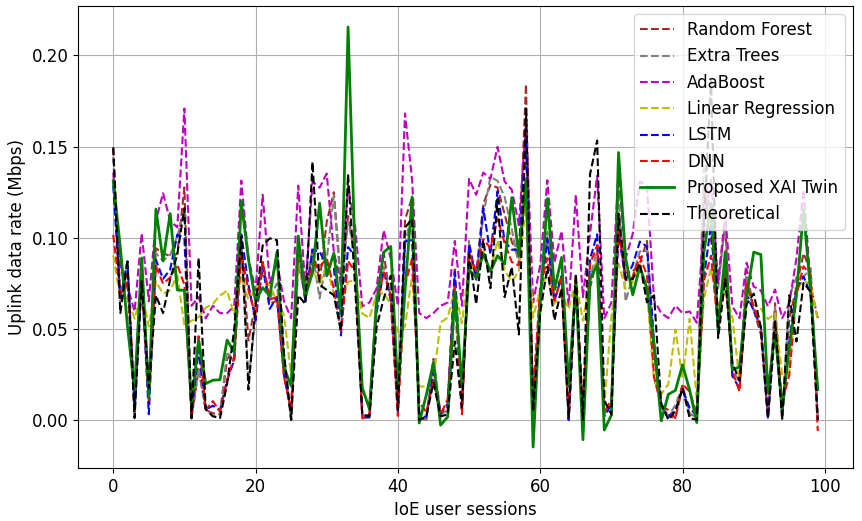}}
		\caption{Comparison of the allocated uplink data rate of the proposed neuro-symbolic XAI twin with other baselines.}
		\label{Uplink_data_rate}
	\end{figure}
	
	\begin{figure}[!t]
		\centerline{\includegraphics[width=8.3cm]{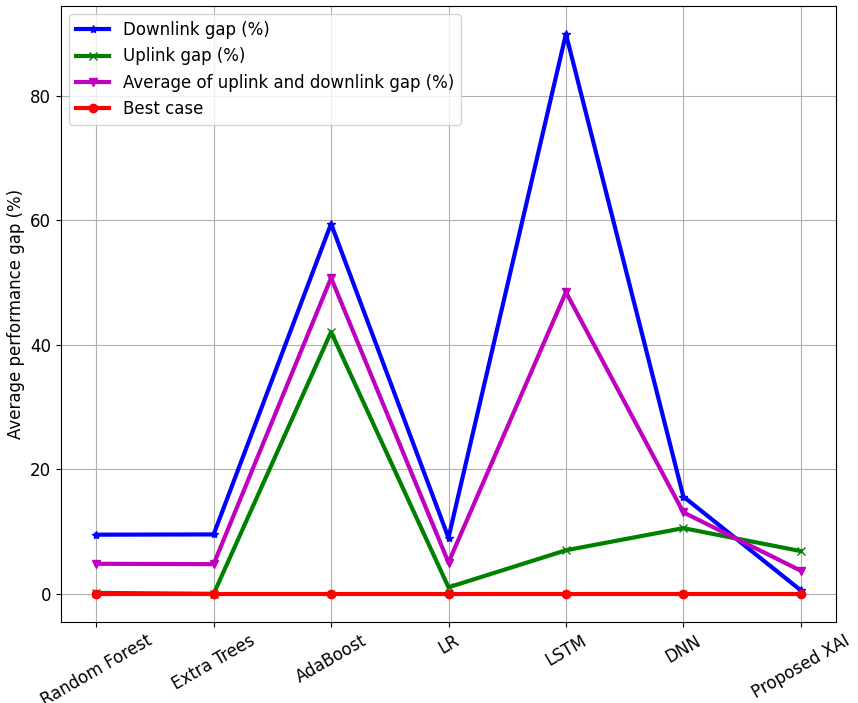}}
		\caption{Performance gap analysis of the proposed neuro-symbolic XAI twin with other baselines.}
		\label{performance_gaps}
	\end{figure}
	
	We first investigate the correctness of the proposed neuro-symbolic XAI twin. We know that, each IoE service requirements are distinct over time. In Figures \ref{Downlink_data_rate} and \ref{Uplink_data_rate}, we show the allocated downlink and uplink data rates to the $100$ IoE sessions. In particular, we compare the accuracy of the both downlink and uplink data rates allocation of the proposed neuro-symbolic XAI twin with Random Forest, Extra Trees, AdaBoost, and Linear Regression, DNN, and LSTM. The Figure \ref{Downlink_data_rate} shows that the maximum downlink error resulting from the proposed method is $0.18$ Mbps while Random Forest, Extra Trees, AdaBoost, and Linear Regression, DNN, and LSTM are $8.05$, $9.97$, $11.91$, $11.63$, $28.16$, and $15.14$ Mbps, respectively. Clearly the proposed neuro-symbolic XAI twin outperforms others baselines in terms of downlink data rate allocation which is near to the theoretical throughput. Further, in case of the uplink data rate allocation to IoE service execution, the maximum error rate by the Random Forest, Extra Trees, AdaBoost, and Linear Regression, DNN, and LSTM yield $0.09$, $0.08$, $0.07$, $0.10$, $0.08$, and $0.1$ Mbps, respectively, where as the proposed neuro-symbolic XAI twin can make a maximum $0.07$ Mbps error during uplink bandwidth allocation to the IoE services as shown in Figure \ref{Uplink_data_rate}. Then, we illustrate a performance gap analysis of the proposed neuro-symbolic XAI twin with other baselines in Figure \ref{performance_gaps}. In this work, we have designed a  multivariate regression for the physical space of the proposed neuro-symbolic XAI twin, thus, we can take both the downlink and uplink data rate decisions at the same iteration of the Algorithm \ref{alg:implicit_learner} at each gNB. As a result, we can minimize the average tradeoff between downlink and uplink bandwidth allocation decisions over other baselines. We analyze such a performance gap in Figure \ref{performance_gaps}, in which, the proposed multivariate regression of the neuro-symbolic XAI twin can achieve better performance than others due to interaction between network dynamics and contextual features during the learning. The average performance gap between the proposed method and the ground truth (theoretical) is $3.74\%$, which is negligible since the environment is non-stationary.

	\begin{figure}[!t]
		\centering
		\subfigure[IoE service association analysis for each gNB.]
		{
			\includegraphics[width=7cm]{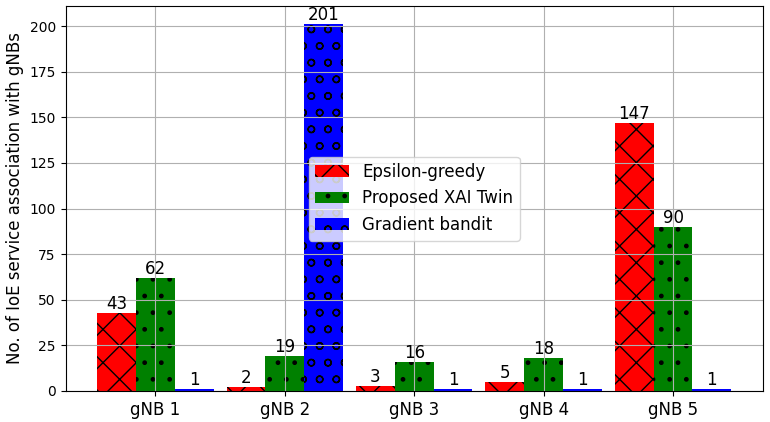}
			\label{IoE_association_gNB}
		}
		\subfigure[Downlink throughput analysis for each gNB.]
		{
			\includegraphics[width=7cm]{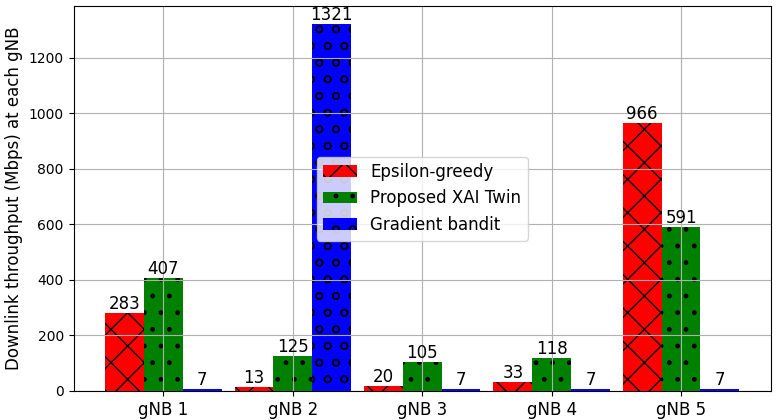}
			\label{Downlink_throughput}
		}
		\subfigure[Uplink throughput analysis for each gNB.]
		{
			\includegraphics[width=7cm]{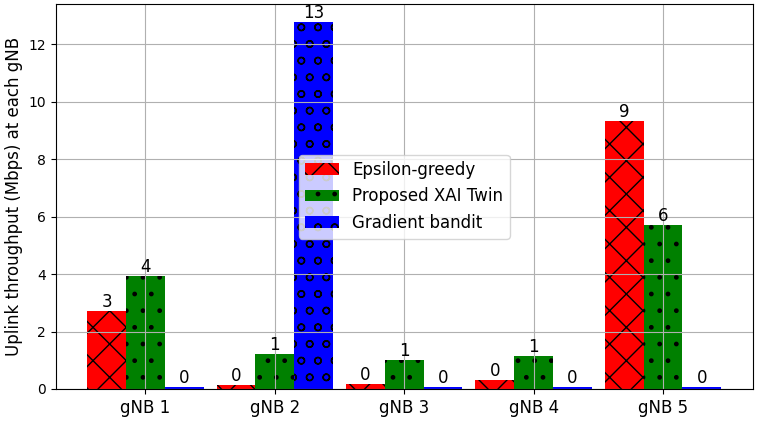}
			\label{Uplink_throughput}
		}
		\caption{Comparison of the gNB association and throughput among the proposed neuro-symbolic XAI twin and other baselines for $205$ IoE sessions.}
		\label{gNB}
	\end{figure}
\begin{figure}[!t]
		\centering
		\subfigure[Marginal probability distribution of evidences of user speed using symbolic reasoning.]
		{
			\includegraphics[width=7cm]{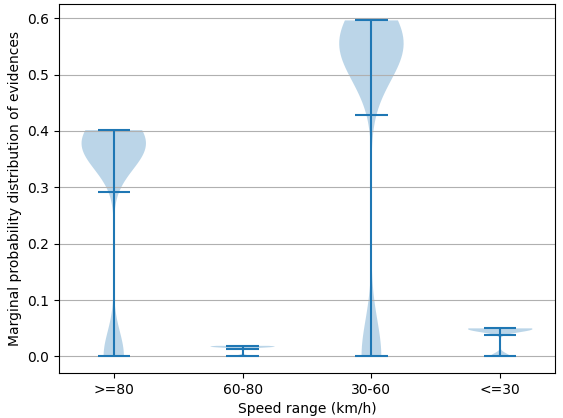}
			\label{Speed_range}
		}
		\subfigure[Marginal probability distribution of evidences of achieved CQI using symbolic reasoning.]
		{
			\includegraphics[width=7cm]{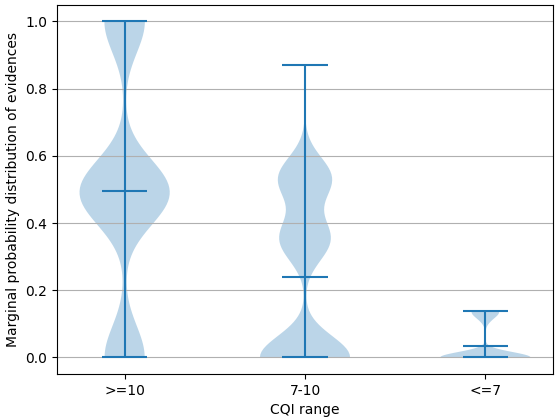}
			\label{CQI_Range}
		}
		\caption{Role of symbolic reasoning on gNB association and bandwidth allocation.}
		\label{symbolic_reasoning}
	\end{figure}
\begin{figure}[!t]
		\centerline{\includegraphics[width=8.3cm]{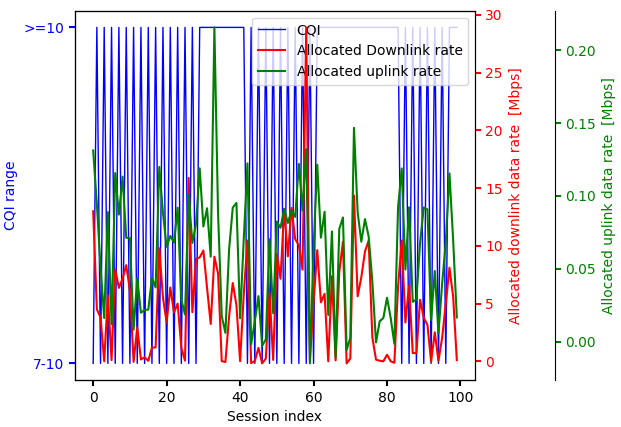}}
		\caption{Achieved uplink and downlink data rates over the gained CQI range by the explainer.}
		\label{CQI_Downlink_Uplink}
	\end{figure}

	Now, we analyze the gNB association of IoE services and their achieved throughput of the proposed neuro-symbolic XAI twin with other baselines such as Gradient bandit, and Epsilon-greedy in Figure \ref{gNB}. In Figures \ref{IoE_association_gNB}, \ref{Downlink_throughput}, and \ref{Uplink_throughput} show the IoE service associations, achieved downlink, and uplink throughput for $205$ IoE session execution. From these Figures, we can clearly see that IoE session association of the proposed neuro-symbolic XAI twin is well-balanced since this scheme can infer the symbolic reasoning among the network dynamics based on marginal probability distribution of evidences by a DAG of BBN. Thus, Figure \ref{symbolic_reasoning} indicates the power of symbolic reasoning on gNB association and bandwidth allocation on IoE service user mobility and achieved CQI. In Figure, \ref{Speed_range}, most of the users' speed belongs to $30-60$ km/h which infers that at the end of a certain duration, the IoE uses moves from gNB $1$ to gNB $5$, thus, the analogies that are shown in Figure \ref{gNB} becomes well justified. The proposed neuro-symbolic XAI twin can also maintain a higher CQI during IoE service execution as shown in Figure \ref{CQI_Range}. We show a relationship between the achieved downlink data rate, uplink data rate, and CQI of the proposed neuro-symbolic XAI twin in Figure \ref{CQI_Downlink_Uplink}.  In particular, $72\%$ cases of IoE service can maintain the highest level of CQI while $28\%$ of IoE achieve CQI in the mid range $7-10$. In summary, the proposed approach is clearly more reliable and effective than the other baselines.

	\subsection{Explainability and Trustworthiness of the Proposed Neuro-symbolic XAI Twin}
	\begin{figure}[!t]
		\centerline{\includegraphics[width=8.3cm]{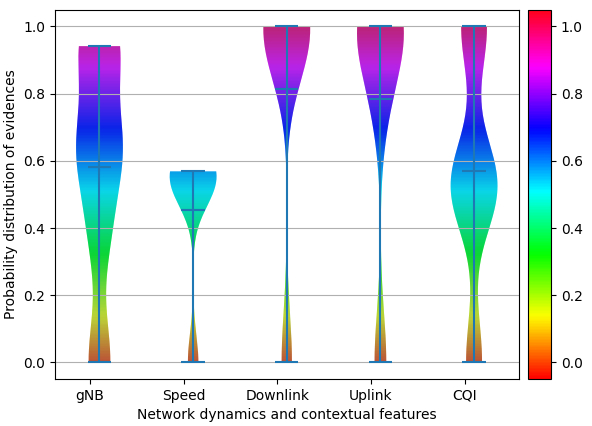}}
		\caption{Statistical explanation based on the maximum marginal joint probability distribution of evidence for each random variable.}
		\label{Max_Probability_distribution_of_evidences}
	\end{figure}
	\begin{figure}[!t]
		\centerline{\includegraphics[width=8.0cm]{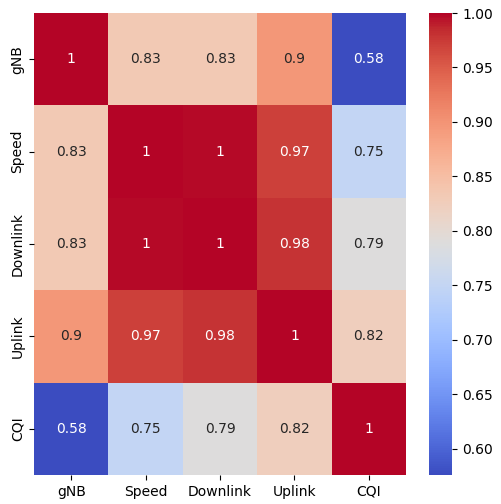}}
		\caption{Correlation among the evidence through BBN-based explainer in virtual space.}
		\label{correlation_evidence}
			\vspace{-4mm}
	\end{figure}
	\begin{figure}[!t]
		\centerline{\includegraphics[width=8.3cm]{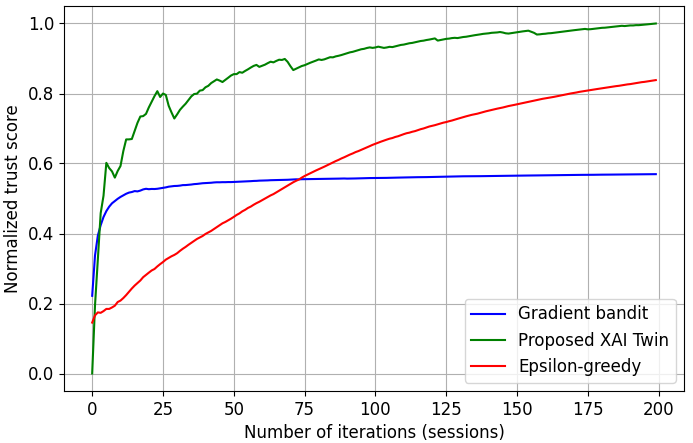}}
		\caption{Trust score by the explicit learner at the virtual space of XAI twin.}
		\label{MAB_Score}
	\end{figure}
	\begin{figure}[!t]
		\centerline{\includegraphics[width=8.3cm]{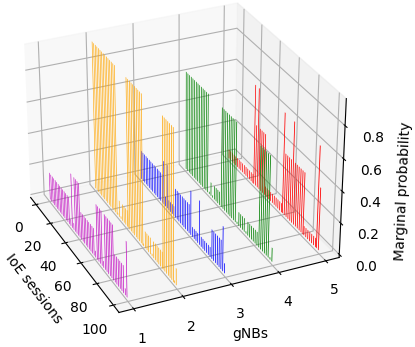}}
		\caption{Marginal trust distribution among the gNBs.}
		\label{Marginal_gNBs}
	\end{figure}
	Figure \ref{Max_Probability_distribution_of_evidences} shows the statistical explanation based on the maximum marginal joint probability distribution of evidence for each random variable of DAG on IoE service properties such gNB, speed, downlink, uplink, and CQI. 
	Figure \ref{Max_Probability_distribution_of_evidences} clearly shows the most prominent explainable evidence by achieving higher probability distribution. Further, we observed a strong correlation over this evidence, as shown in Figure \ref{correlation_evidence}.
	The proposed symbolic reasoning approach found that each gNB association decision has at least $83\%$, $83\%$, $90\%$, and $58\%$ positive correlation over IoE service user's mobility, downlink, uplink, and CQI, respectively. Further, mobility and uplink data rates have a significant impact on downlink data rate decisions. Particularly, $100\%$ and $98\%$ dependencies with mobility and uplink data rate on downlink data rate, respectively.
	The Figure \ref{correlation_evidence} shows that the CQI of IoE services completely relies on the user's mobility, downlink, and uplink data rate with a correlation of $75\%$, $79\%$, and $82\%$, respectively. To this end, the proposed neuro-symbolic XAI twin not only has a strong symbolic reasoning ability but also can capture the fundamental principle of IoE such as users' behavior, data, processes, and physical things.

	Next, we concentrate on the analogy of the trustworthiness of the proposed neuro-symbolic XAI twin in Figure \ref{MAB_Score} since one of the main goals of this work is to provide a trustworthy solution for assuring ZSM. In particular, in Figure \ref{MAB_Score}, we illustrate a comparison based on the achieved normalized trust score for the learning objective \eqref{eq:Bay_fn_regret} among the proposed scheme and other baselines such Gradient-based bandits and Epsilon-greedy \cite{IEEEhowto:MAB_Book_1}. The Figure \ref{MAB_Score} shows that we can trust the proposed neuro-symbolic XAI twin around $44\%$ and $18\%$ more than that of the Gradient-based bandits and Epsilon-greedy, respectively, in order to zero-touch network and service management of IoE. Finally, Figure \ref{Marginal_gNBs} demonstrated the marginal trust distribution among the gNBs of $100$ IoE service fulfillment by the proposed neuro-symbolic XAI twin. The proposed neuro-symbolic XAI twin can adapt to the dynamic environment and employ self-adaptive closed-loop decisions for IoE service execution. Moreover, the DAG-based explainer is more powerful in terms of reasoning and the Bayesian multi-arm bandit scheme assures trustworthiness compared to the baselines.

	\section{Conclusion}
	\label{Conclusion}
	In this paper, we have proposed a new neuro-symbolic XAI twin framework to enable trustworthy and zero-touch IoE service management in wireless networks. To guarantee evidence-based reasoning on network parameters and contextual metrics, we devise a multivariate regression for capturing an unconscious decision while posing with declarative semantics through a directed acyclic graph-based Bayesian network. We have formulated Bayesian multi-arm bandits learning problem to train our proposed neuro-symbolic XAI twin framework, where the learning objective is to minimize the gap between expected and current explainable scores. We have implemented our neuro-symbolic XAI twin framework by developing duel-learning systems that include an implicit learner to take an unconscious decision in the physical space of the XAI twin by solving a multivariate regression problem and an explicit leaner that exploits symbolic reasoning on implicit learner decisions and evidence via prior knowledge. The experiment result shows that the proposed solution significantly outperforms the baselines Gradient-based bandits, Epsilon-greedy, regression models such as ensemble, LSTM, and DNN in terms of accuracy, explainable score, and closed-loop IoE service execution.

	\begin{IEEEbiography}[{\includegraphics[width=1in,height=1.25in,clip,keepaspectratio]{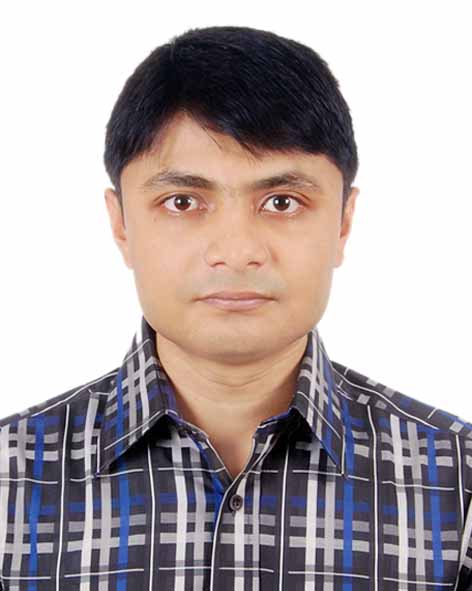}}]{Md.~Shirajum~Munir}
		(S'19-M'21) received the B.S. degree in computer science and engineering from Khulna University, Khulna, Bangladesh, in 2010, and the Ph.D. degree in computer engineering from Kyung Hee University (KHU), South Korea, in 2021. He is currently working as a Post-doctoral Research Associate at Virginia Modeling, Analysis, and Simulation Center, Department of Computational Modeling and Simulation Engineering, Old Dominion University, Suffolk, VA 23435, USA. He served as a Postdoctoral Researcher form September 2021 to August 2022  with the department of computer science and engineering, Kyung Hee University (KHU), South Korea. Before his Ph.D., he also served as a Software Engineer, Senior Software Engineer, and Lead Engineer with the Solution Laboratory, Samsung Research and Development Institute, Dhaka, Bangladesh, from 2010 to 2016. His current research interests include intelligent IoT network management, sustainable edge computing, intelligent healthcare systems, smart grid, trustworthy artificial intelligence, and machine learning.
	\end{IEEEbiography}
	\begin{IEEEbiography}[{\includegraphics[width=1in,height=1.25in,clip,keepaspectratio]{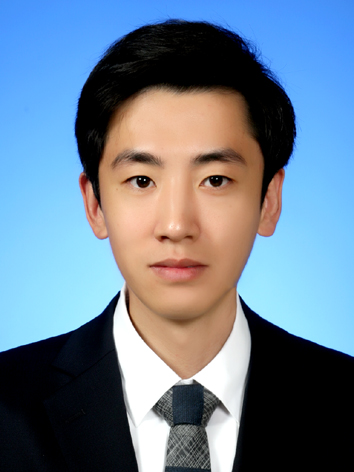}}]{Ki~Tae~Kim}
		received the B.S. and M.S. degrees in computer science and engineering from Kyung Hee University, Seoul, South Korea, in 2017 and 2019, respectively, where he is currently pursuing the Ph.D. degree in computer science and engineering. His research interests include SDN/NFV, wireless networks, unmanned aerial vehicle communications, and machine learning.
	\end{IEEEbiography}

	\begin{IEEEbiography}[{\includegraphics[width=1in,height=1.25in,clip,keepaspectratio]{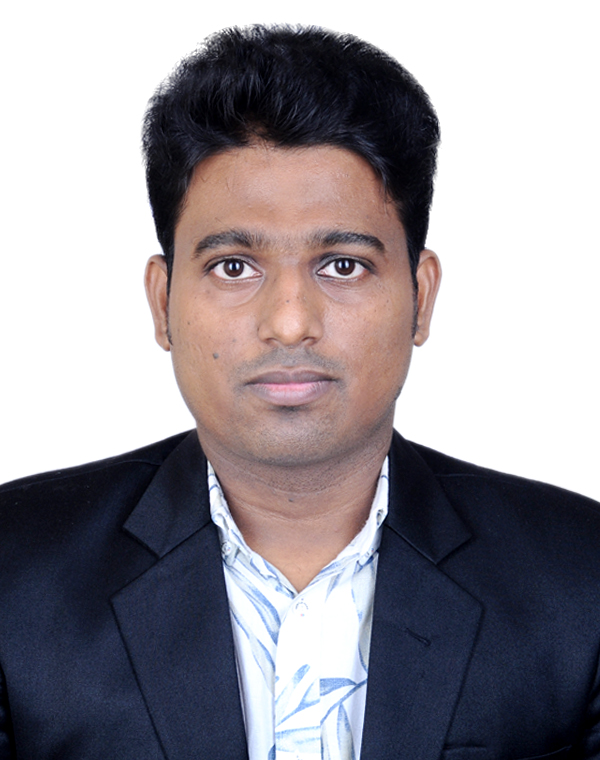}}]{Apurba~Adhikary}
	received his B.Sc and M.Sc Engineering degrees in Electronics and Communication Engineering from Khulna University, Khulna, Bangladesh. He is a PhD Researcher at the Department of Computer Science and Engineering at Kyung Hee University, South Korea. He has been serving as an Assistant Professor at Information and Communication Engineering department in Noakhali Science and Technology University (NSTU), Noakhali, Bangladesh since January 2020. He His research interests are currently focused on distributed edge intelligence, target oriented communication and computation, and intelligent networking resource management.
	\end{IEEEbiography}

	\begin{IEEEbiography}[{\includegraphics[width=1in,height=1.25in,clip,keepaspectratio]{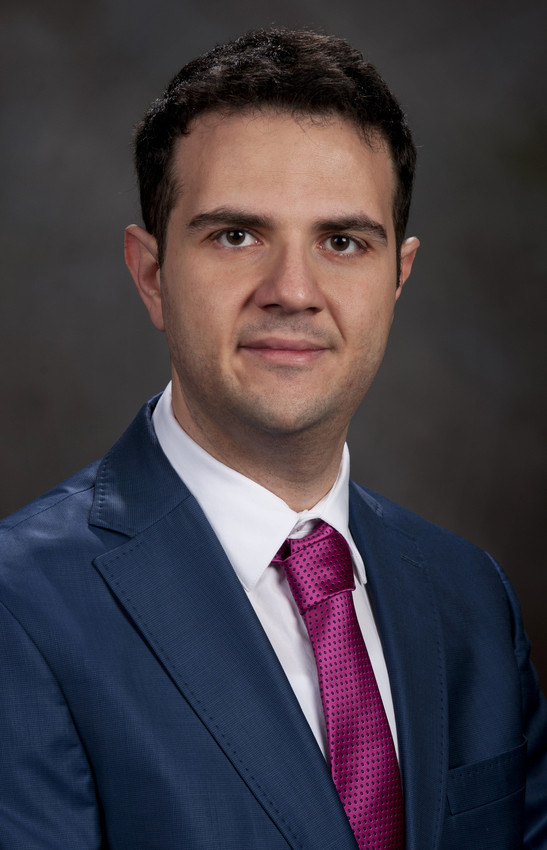}}]{Walid~Saad} (S'07, M'10, SM'15, F'19)
	received his Ph.D degree from the University of Oslo in 2010. Currently, he is a full Professor at the Department of Electrical and Computer Engineering at Virginia Tech, where he leads the Network sciEnce, Wireless, and Security (NEWS@VT) laboratory, within the Wireless@VT research group. His research interests include wireless networks, machine learning, game theory, security, unmanned aerial vehicles, cyber-physical systems, smart grids, and network science. Dr. Saad is a Fellow of the IEEE and an IEEE Distinguished Lecturer. He is also the recipient of the NSF CAREER award in 2013, the AFOSR summer faculty fellowship in 2014, and the Young Investigator Award from the Office of Naval Research (ONR) in 2015. He was the author/co-author of ten conference best paper awards at WiOpt in 2009, ICIMP in 2010, IEEE WCNC in 2012, IEEE PIMRC in 2015, IEEE SmartGridComm in 2015, EuCNC in 2017, IEEE GLOBECOM in 2018, IFIP NTMS in 2019, IEEE ICC in 2020, and IEEE GLOBECOM in 2020. He is the recipient of the 2015 Fred W. Ellersick Prize from the IEEE Communications Society, of the 2017 IEEE ComSoc Best Young Professional in Academia award, of the 2018 IEEE ComSoc Radio Communications Committee Early Achievement Award, and of the 2019 IEEE ComSoc Communication Theory Technical Committee. He was also a co-author of the 2019 IEEE Communications Society Young Author Best Paper. From 2015-2017, Dr. Saad was named the Stephen O. Lane Junior Faculty Fellow at Virginia Tech and, in 2017, he was named College of Engineering Faculty Fellow. He received the Dean's award for Research Excellence from Virginia Tech in 2019. He currently serves as an editor for the IEEE Transactions on Mobile Computing and the IEEE Transactions on Cognitive Communications and Networking. He is an Editor-at-Large for the IEEE Transactions on Communications.
\end{IEEEbiography}

\begin{IEEEbiography}[{\includegraphics[width=1in,height=1.25in,clip,keepaspectratio]{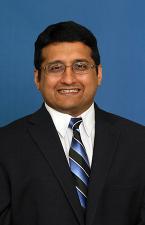}}]{Sachin~Shetty}
Sachin Shetty is an Executive Director for Center of Secure and Intelligent Critical Systems in the Virginia Modeling, Analysis and Simulation Center at Old Dominion University. He holds a joint appointment as a Professor with the Department of Computational, Modeling and Simulation Engineering  . Sachin Shetty received his PhD in Modeling and Simulation from the Old Dominion University in 2007 under the supervision of Prof. Min Song.   Prior to joining Old Dominion University, he was an Associate Professor with the Electrical and Computer Engineering Department at Tennessee State University. He was also the associate director of the Tennessee Interdisciplinary Graduate Engineering Research Institute and directed the Cyber Security laboratory at Tennessee State University. He also holds a dual appointment as an Engineer at the Naval Surface Warfare Center, Crane Indiana. His research interests lie at the intersection of computer networking, network security and machine learning. His laboratory conducts cloud and mobile security research and has received over \$12 million in funding from National Science Foundation, Air Office of Scientific Research, Air Force Research Lab, Office of Naval Research, Department of Homeland Security, and Boeing. He is the site lead on the DoD Cyber Security Center of Excellence, the Department of Homeland Security National Center of Excellence, the Critical Infrastructure Resilience Institute (CIRI), and Department of Energy, Cyber Resilient Energy Delivery Consortium (CREDC). He has authored and coauthored over 140 research articles in journals and conference proceedings and two books. He is the recipient of Fulbright Specialist award, EPRI Cybersecurity Research Challenge award, CCI Fellow, DHS Scientific Leadership Award and has been inducted in Tennessee State University’s million dollar club. He has served on the technical program committee for ACM CCS, IEEE INFOCOM, IEEE ICDCN, and IEEE ICCCN. He is a Senior Member of IEEE.
\end{IEEEbiography}

\begin{IEEEbiography}[{\includegraphics[width=1in,height=1.25in,clip,keepaspectratio]{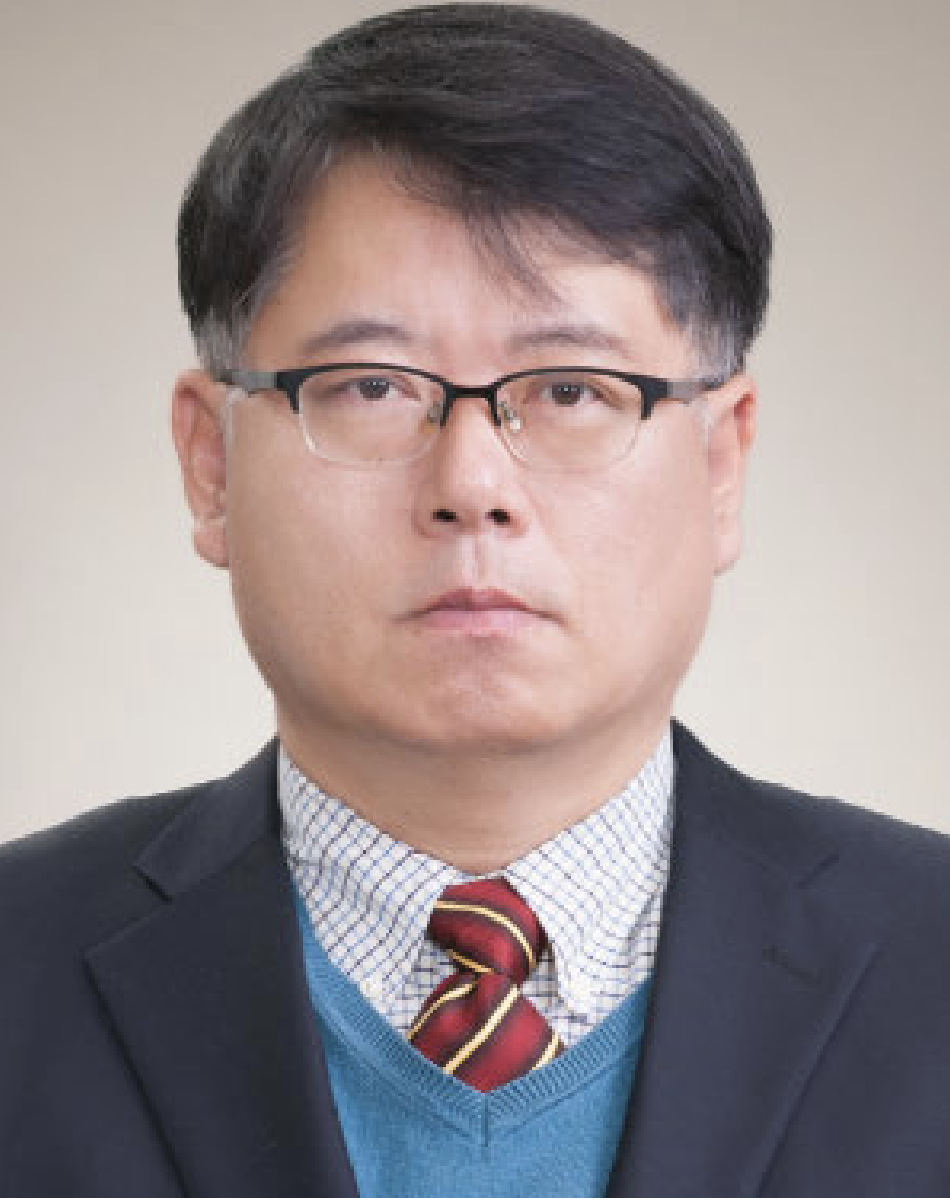}}]{Seong-Bae Park}
received his B.S. degree in computer science from Korea Advanced Institute of
Science and Technology in 1994, and the M.S. degree in computer engineering and the Ph.D. degree
in computer science and engineering from Seoul National University in 1996 and 2002, respectively.
He was a professor of computer science and engineering with Kyungpook National University from
2004 to 2017. In 2018, he joined Kyung Hee University, where he is currently a full professor of
computer science and engineering. His research interests include machine learning, natural language
processing, text mining, information extraction, and bio-informatics.
\end{IEEEbiography}

\begin{IEEEbiography}[{\includegraphics[width=1in,height=1.25in,clip,keepaspectratio]{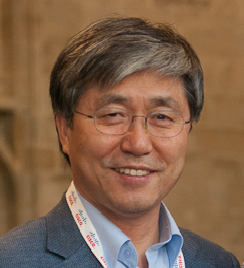}}]{Choong~Seon~Hong}	(S'95-M'97-SM'11)
received the B.S. and M.S. degrees in electronic engineering from Kyung Hee University,	Seoul, South Korea, in 1983 and 1985, respectively, and the Ph.D. degree from Keio University, Tokyo, Japan, in 1997. In 1988, he joined KT, Gyeonggi-do, South Korea, where he was involved in broadband networks as a member of the Technical Staff. Since 1993, he has been with Keio University. He was with the Telecommunications Network Laboratory,	KT, as a Senior Member of Technical Staff and as the Director of the Networking Research Team until 1999. Since 1999, he has been a Professor with the Department of Computer Science and Engineering, Kyung Hee University. His research interests include future Internet, intelligent edge computing, network management, and network security. Dr. Hong is a member of the Association for Computing Machinery (ACM), the	Institute of Electronics, Information and Communication Engineers (IEICE), the Information Processing Society of Japan (IPSJ), the Korean Institute of Information Scientists and Engineers (KIISE), the Korean Institute of Communications and Information Sciences (KICS), the Korean Information	Processing Society (KIPS), and the Open Standards and ICT Association (OSIA). He has served as the General Chair, the TPC Chair/Member, or an Organizing Committee Member of international conferences, such as the	Network Operations and Management Symposium (NOMS), International Symposium on Integrated Network Management (IM), Asia-Pacific Network Operations and Management Symposium (APNOMS), End-to-End Monitoring Techniques and Services (E2EMON), IEEE Consumer Communications and Networking Conference (CCNC), Assurance in Distributed Systems	and Networks (ADSN), International Conference on Parallel Processing (ICPP), Data Integration and Mining (DIM), World Conference on Information Security Applications (WISA), Broadband Convergence Network (BcN), Telecommunication Information Networking Architecture (TINA), International Symposium on Applications and the Internet (SAINT), and	International Conference on Information Networking (ICOIN). He was an Associate Editor of the IEEE TRANSACTIONS ON NETWORK AND SERVICE MANAGEMENT and the IEEE JOURNAL OF COMMUNICATIONS AND NETWORKS and an Associate Editor for the International Journal of Network Management and an Associate Technical Editor of the IEEE	Communications Magazine. He currently serves as an Associate Editor for the International Journal of Network Management and Future Internet Journal.
\end{IEEEbiography}

\end{document}